\documentclass[letterpaper]{article} 
\usepackage{aaai2026}  
\usepackage{times}  
\usepackage{helvet}  
\usepackage{courier}  
\usepackage[hyphens]{url}  
\usepackage{graphicx} 
\usepackage{natbib}  
\usepackage{caption} 
\usepackage{amsmath}
\usepackage{amsfonts}
\usepackage{booktabs}
\usepackage{subcaption}
\usepackage{multirow}
\usepackage{tikz}
\usetikzlibrary{bayesnet}
\usepackage{longtable}
\usepackage{algorithm}
\usepackage{algorithmic}

\title{VITA: Variational Pretraining of Transformers for Climate-Robust Crop Yield Forecasting}

\author{
    Adib Hasan\textsuperscript{\rm 1},
    Mardavij Roozbehani\textsuperscript{\rm 2},
    Munther Dahleh\textsuperscript{\rm 2}
}
\affiliations{
    \textsuperscript{\rm 1}Independent Researcher\\
    \textsuperscript{\rm 2}Massachusetts Institute of Technology\\
    notadib@alum.mit.edu, mardavij@mit.edu, dahleh@mit.edu
}

\begin{document}

\maketitle

\begin{abstract}
    Accurate crop yield forecasting is essential for global food security. However, current AI models systematically underperform when yields deviate from historical trends. We attribute this to the lack of rich, physically grounded datasets directly linking atmospheric states to yields. To address this, we introduce \textit{VITA (\textbf{V}ariational \textbf{I}nference \textbf{T}ransformer for \textbf{A}symmetric data)}, a variational pretraining framework that learns representations from large satellite-based weather datasets and transfers to the ground-based limited measurements available for yield prediction. VITA is trained using detailed meteorological variables as proxy targets during pretraining and learns to predict latent atmospheric states under a seasonality-aware sinusoidal prior. This allows the model to be fine-tuned using limited weather statistics during deployment. Applied to 763 counties in the U.S. Corn Belt, VITA achieves state-of-the-art performance in predicting corn and soybean yields across all evaluation scenarios, particularly during extreme years, with statistically significant improvements (paired t-test, $p < 0.0001$). Importantly, VITA outperforms prior frameworks like GNN-RNN without soil data, and larger foundational models (e.g., Chronos-Bolt) with less compute, making it practical for real-world use—especially in data-scarce regions. This work highlights how domain-aware AI design can overcome data limitations and support resilient agricultural forecasting in a changing climate.
    \end{abstract}

\begin{links}
\link{Code}{https://github.com/neehan/VITA}
\end{links}

\section{Introduction}
\begin{figure*}[hbt]
\centering
\begin{subfigure}{0.49\textwidth}
    \centering
    \includegraphics[width=\textwidth]{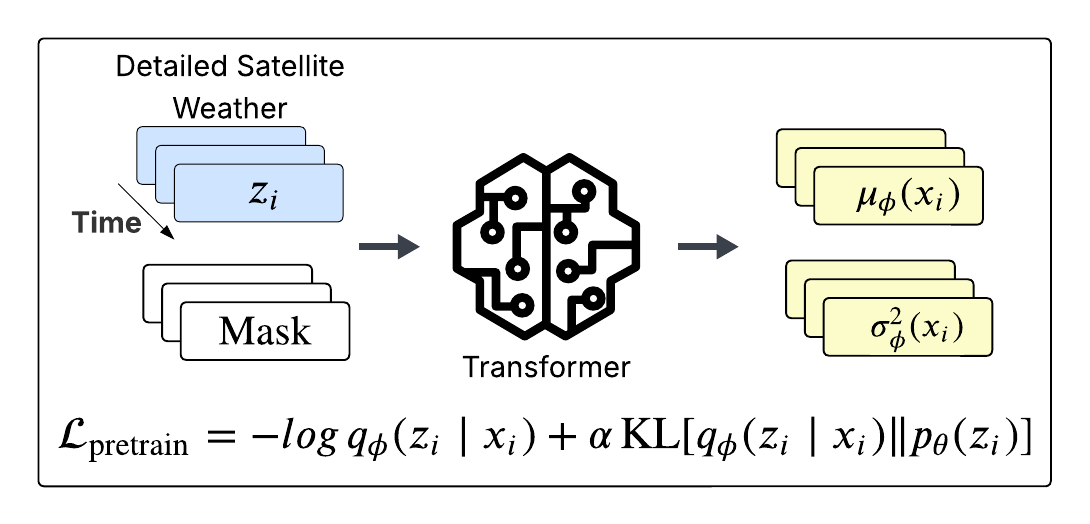}
    \caption{Variational pretraining}
    \label{fig:pretrain}
\end{subfigure}
\hfill
\begin{subfigure}{0.50\textwidth}
    \centering
    \includegraphics[width=\textwidth]{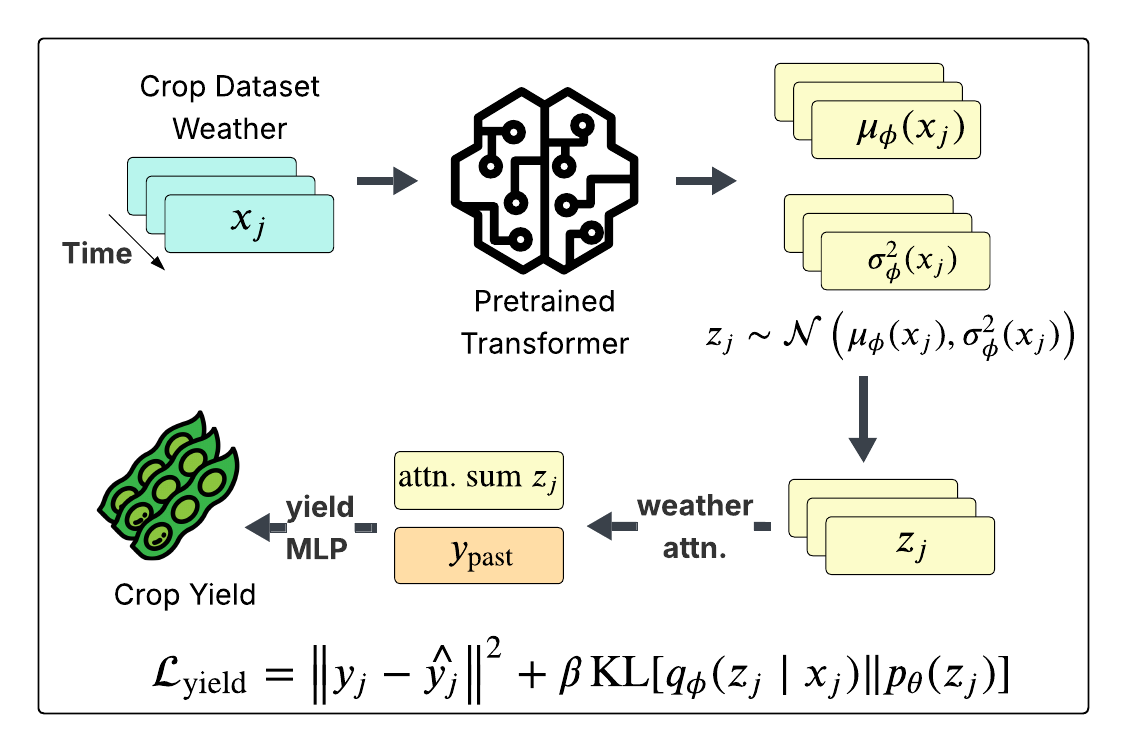}
    \caption{Yield prediction fine-tuning}
    \label{fig:finetune}
\end{subfigure}
\caption{Two-stage variational training framework for asymmetric weather features. (a) A transformer encoder is pretrained on 31-variable weather time series by randomly masking $10 \leq k \leq 25$ features and predicting them from remaining context. The model learns a variational posterior $q_\phi(z_i \mid x_i)$ over weather representations by directly maximizing variational likelihood. (b) During fine-tuning, only 6 weather features are available. The pretrained transformer encodes these into a latent distribution $q_\phi(z_j \mid x_j)$, from which $z_j \sim \mathcal{N}(\mu_\phi(x_j), \sigma_\phi^2(x_j))$ is sampled. It is aggregated with learnable attention across time dimension and concatenated with historical yield $y_{\text{past}}$ for final yield prediction.}
\label{fig:big-picture}
\end{figure*}

Climate change is transforming agriculture, with extreme weather causing billions in annual crop losses \citep{lobell2011climate}. In 2012, U.S. drought reduced corn yields by 13\%, while 2019 flooding prevented planting on 19.4 million acres \citep{usda2013drought, usda2019prevented}. Accurate yield prediction under such volatility is critical for agricultural risk management and long-term food security \citep{Beddington2010}. Yet current operational models—including regression-based approaches from USDA ERS \citep{WestcottJewison2013}—often fail when yields diverge from historical trends.

The challenge stems from fundamental data limitations of existing methods in yield forecasting. First, many models train on less than 10 years of historical data \citep{gandhi2016, cropnet2024}, insufficient for capturing rare but increasingly critical extreme weather patterns. Second, multi-modal approaches \citep{you2017deep, cropnet2024, Khaki2019, gnnrnn2021} rely on extensive auxiliary data—satellite imagery, soil surveys, planting records, and weather records—which limits their applicability in regions that lack detailed agricultural monitoring infrastructure.

Third, general-purpose time-series pretraining methods like SimMTM \citep{dong2023simmtmsimplepretrainingframework}, PatchTST \citep{wu2023patchtst}, and foundational models, such as Chronos \citep{ansari2024chronos} assume consistent input features between pretraining and fine-tuning. This is suboptimal in weather domains, where pretraining can leverage rich satellite datasets with dozens of variables (e.g., 31 meteorological variables from NASA POWER \citep{NASAPower}), but fine-tuning must rely on smaller, accessible subsets (e.g., 6 basic weather variables from \citet{Khaki2019}). We term this the \textit{data asymmetry problem}, in which pretraining and operational feature sets fundamentally differ—an issue largely unaddressed by existing AI approaches.

To overcome this limitation, we introduce VITA, a variational pretraining framework for weather–yield prediction that transfers knowledge from rich satellite data to limited ground-based inputs. Trained with a variational loss and feature mask, VITA learns latent weather representations that generalize to settings with fewer variables. Since many crop yields are largely governed by weather, this objective enables rich representation learning essential for yield forecasting.

Unlike a standard variational autoencoder (VAE) \citep{kingma2013auto}, VITA is trained without a decoder -- it maximizes the variational likelihood using detailed weather variables as proxy targets, with a sinusoidal prior to capture seasonality. During fine-tuning, six years of historical yield serve as a proxy for soil and management factors, reducing data requirements. In total, VITA adds under 2\% more parameters than a standard Transformer encoder \citep{vaswani2017} and trains end-to-end in under 2.5 hours on a single L40S GPU. Its efficiency and exclusive reliance on public data (NASA POWER, USDA yield) make it practical for operational use, including crop insurance and subsidy programs.

Empirically, VITA delivers state-of-the-art accuracy, particularly in years with extreme yield deviations, with statistically significant gains ($p<0.0001$) over other pretraining strategies. It also generalizes strongly across space and time—models pretrained on non-U.S. weather improve U.S. fine-tuning, and those trained on 1994–2009 remain accurate for 2014–2018—demonstrating robust, transferable weather representations.

\medskip

In summary, the key contributions of this work are:

\begin{itemize}

\item A decoder-free variational pre-training framework for modeling asymmetric features (see Equation~\ref{eq:implemented_yield_loss}).

\item A seasonality-aware sinusoidal prior that captures structured temporal patterns.

\item State-of-the-art performance in years with extreme yield deviations, validated through rigorous, statistically grounded evaluation.

\end{itemize}

The remainder of this paper is organized as follows: Related Work reviews prior methods; Methodology details our variational framework; Experiments presents our experimental setup; Results analyzes performance across various conditions; and Discussion examines implications and limitations.

\section{Related Work}

\paragraph{Crop Yield Prediction.} Researchers have proposed various different approaches for crop yield forecasting, including mechanistic modeling, CNN-RNN architectures, graph neural networks, Deep Gaussian Processes, and Vision Transformers \citep{apsim2003, Khaki2019, gnnrnn2021, you2017deep, cropnet2024}. Multi-modal approaches integrate satellite imagery, weather data, soil surveys, management records, and vegetation indices \citep{rs13091735, sun2019county, gandhi2016, oliveira2018scalablemachinelearningpreseason, BASIR2021100186, hasan2024weatherformer, FERRAZ2024100661, cao2022cornyieldpredictionbased}. However, extensive data requirements limit deployment in regions with limited agricultural monitoring infrastructure, and many approaches rely on temporally short regional datasets that may not capture sufficient weather variability for extreme events \citep{gandhi2016, cropnet2024, McFarland2020g2f, CHU2020105471}.

\paragraph{Time Series Pretraining.} Recent work develops unsupervised representation learning through contrastive methods, transformer frameworks, and masked reconstruction \citep{franceschi2019unsupervised, yue2022ts2vec, zerveas2021transformer, dong2023simmtmsimplepretrainingframework, wu2023patchtst, woo2022costcontrastivelearningdisentangled}. These methods achieve strong forecasting performance but do not explicitly model data asymmetry, which is critical for weather-based yield forecasting.

\paragraph{Variational Methods.} Variational autoencoders have been applied to weather prediction, climate forecasting, and agricultural data generation \citep{kingma2013auto, higgins2017betavae, DBLP:journals/corr/abs-2111-03476, wang2024accurate, palma2025datadrivenseasonalclimatepredictions, RAZAVI202499}. However, existing variational methods are not used for asymmetric data learning. Our approach is tailored to this problem, learning latent states without input reconstruction and using a seasonality-aware sinusoidal prior.

\section{Methodology}
VITA incorporates a two-stage approach: (1) self-supervised pretraining on extensive weather data to learn robust weather representations, and (2) variational fine-tuning with basic weather statistics and past yields.

\subsection{Problem Formulation}

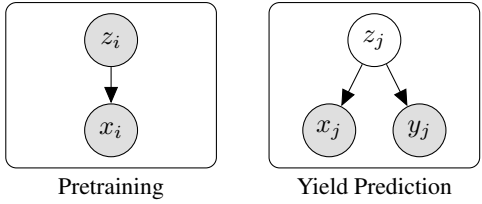
\begin{figure}[htb]
\centering
\begin{tikzpicture}

  \node[obs] (zi) at (0, 1.8) {$z_i$};
  \node[obs] (xi) at (0, 0.6) {$x_i$};
  \edge {zi} {xi};
  \node[draw=black, rounded corners, minimum width=2.8cm, minimum height=2.2cm, label=below:{Pretraining}] at (0, 1.2) {};

  \node[latent] (zj) at (3.5, 1.8) {$z_j$};
  \node[obs] (xj) at (2.9, 0.6) {$x_j$};
  \node[obs] (yj) at (4.1, 0.6) {$y_j$};
  \edge {zj} {xj};
  \edge {zj} {yj};
  \node[draw=black, rounded corners, minimum width=2.8cm, minimum height=2.2cm, label=below:{Yield Prediction}] at (3.5, 1.2) {};

\end{tikzpicture}
\caption{{\small Graphical model showing the data structure of pretraining and prediction phases in VITA.}}
\label{fig:graphical_model}
\end{figure}

\noindent We formulate crop yield prediction as semi-supervised learning with latent weather representations. Suppose $z \in \mathbb{R}^{364 \times 31}$ denotes the detailed meteorological variables from pretraining, and $x \in \mathbb{R}^{364 \times 6}$ denotes basic weather statistics (temperature, precipitation, etc.) available in both pretraining and downstream tasks. Each weather input is a 364-week sequence representing seven years of weekly means.

We consider two datasets: (1) a pretraining dataset $\mathcal{D}_w = \{(x_i, z_i)\}_{i=1}^{N_w}$, where both basic and detailed weather states are observed for each 364-week \textit{non-overlapping} window $i$ across the NASA POWER grid, but no yield information is available; and (2) a finetuning dataset $\mathcal{D}_y = \{(x_j, y_j, y_{\text{past},j})\}_{j=1}^{N_y}$, where detailed states $z_j$ remain latent for each 364-week \textit{overlapping} sequence $j$ across U.S. Corn Belt counties. Each training example uses weather from years $[t-6, t]$ and historical yields from $[t-6, t-1]$ to predict  yield $y_{c,t}$ for county $c$. Test-year yields are strictly held out during training and additional details are provided in the Appendix. We model each county independently as a 1D temporal sequence and spatial context enters only via (latitude,longitude) features concatenated to each timestep (Equation~\ref{eq:input_concat}).

\subsection{Architecture}
VITA employs a transformer encoder that maps 364-week weather sequences into latent representations for yield prediction. The forward process is:

\begin{align}
x_{\text{input}} &= \mathrm{concat}(x_{\text{weather}}, \text{year}, \text{coordinates}) \label{eq:input_concat}\\
h_{\text{weather}} &= E_\phi\!\big(\mathrm{LinearProj}(x_{\text{input}}) + \mathrm{PosEmbed}(\cdot)\big) \\
[\mu, \log\sigma^2] &= \mathrm{LinearProj}_{\mu,\sigma^2}(h_{\text{weather}}) \\
z &= \mu + \sigma \odot \epsilon, \quad \epsilon \sim \mathcal{N}(0, I) \\
z_{\text{agg}} &= \sum_{k=1}^{364} \alpha_k z_k;\; 
\alpha_k = \mathrm{softmax}\!\big(\mathrm{MLP}_a(z_{k})\big) \\
\hat{y}_t &= \mathrm{MLP}_y\!\big([\,z_{\text{agg}},\, y_{\text{past}}\,]\big)
\end{align}

Here, $E_\phi$ is a transformer encoder applied to weekly weather inputs $x_{\text{input}}$ with positional embeddings, and coordinates refer to the spatial latitude and longitude. 
The LinearProj$_{\mu,\sigma^2}$ outputs the mean and log-variance of each latent state; MLP$_a$ computes attention weights for temporal aggregation into $z_{\text{agg}}$; and MLP$_y$ maps the aggregated latent representation and historical yields to the predicted yield $\hat{y}_t$.

\subsection{Self-Supervised Pretraining}

We pretrain on the NASA POWER dataset (1984-2022) using progressive feature-wise masking, starting with $k=10$ masked features and increasing by 1 every 2 epochs until 25 out of 31 features are masked. The pretraining objective balances reconstruction with regularization:
\begin{align}
    \mathcal{L}_{\text{pretrain}} = 
    &-\mathbb{E}_{(x_i, z_i) \sim \mathcal{D}_w}\left[\log q_\phi
(z_i \mid x_i)\right. \notag \\
&+ \left.\alpha \cdot \text{KL}[q_\phi(z_i \mid x_i) \| p_\theta
(z_i)]\right] \notag \\
    =&-\mathbb{E}_{(x_i, z_i) \sim \mathcal{D}_w}\left[\log \mathcal{N}(z_i; \mu_\phi(x_i), \sigma^2_\phi(x_i))\right] \notag \\
&+ \alpha \cdot \text{KL}[q_\phi(z_i \mid x_i) \| p_\theta(z_i)]
\label{eq:main_pretrain_loss}
\end{align}
The first term maximizes the Gaussian likelihood, which encourages the posterior distribution $q_\phi(z_i \mid x_i)$ to accurately predict the observed detailed weather state $z_i$. The second term is a regularizer preventing overfitting by imposing a prior structure.

We investigate two prior distributions: standard normal $p_\theta(z) \sim \mathcal{N}(0, I)$ and sinusoidal prior $p_\theta(z) \sim \mathcal{N}(\mu_{\sin}=A \sin(\theta \cdot \text{pos} + \theta_0), \sigma^2 I)$ to capture seasonal patterns. The Sinusoidal prior has additional prior parameters that are learned during both pretraining and finetuning. This prior explicitly models the periodicity in weather variables, allowing a more structured latent space.

\subsection{Decoder-Free Fine-Tuning Objective}
Standard semi-supervised VAEs require a decoder term $\log p(x_j|z_j)$ to model input reconstruction \citep{kingma2014semi}. However, in our meteorological context, established principles like the Tetens equation, Penman-Monteith formulation, Clausius-Clapeyron equation and Stefan-Boltzmann radiation balance link basic weather statistics to the detailed atmospheric state \citep{Tetens1930, Ndulue2021, brown1951clausius, MurrayTortarolo2023}, making the decoder term unnecessary.

In the Appendix, we empirically validated this deterministic relationship, training a small MLP to reconstruct basic weather variables from detailed ones with near-perfect accuracy ($R^2 > 0.9999$). This enables us to model $p(x_j|z_j) \approx 1$ and derive the simplified variational objective:
\begin{equation}
\mathcal{L}_{\text{yield}} = \|y_j - \hat{y}_j\|^2 + \beta \cdot \text{KL}[q_\phi(z_j \mid x_j) \| p_\theta(z_j)] \label{eq:implemented_yield_loss}
\end{equation}
where $\beta>0$ is a hyperparameter. Note that the $\beta$ in Equation~(\ref{eq:implemented_yield_loss}) does not weaken the variational objective and the full evidence lower bound (ELBO) is still optimized. The full derivation of this objective is shown in the Appendix.

\subsection{Baselines}
We compare against several types of baselines. (1) \textit{Non-deep learning methods:} OLS linear regression with agronomically-motivated features following USDA Economic Research Service (ERS) methodology \citep{WestcottJewison2013}, and XGBoost \citep{Chen_2016}. (2) \textit{Deep learning methods:} CNN-RNN \citep{Khaki2019} and GNN-RNN \citep{gnnrnn2021}. (3) \textit{Masked time series pretraining:} SimMTM \citep{dong2023simmtmsimplepretrainingframework}. (4) \textit{Pre-trained foundational time series models:} Chronos-Bolt-tiny-9M \citep{ansari2024chronos} with full fine-tuning. To isolate the effect of variational pretraining from architecture, we also include T-BERT, which is identical to VITA but trained with a standard MSE reconstruction loss. Notably, XGBoost, GNN-RNN, and CNN-RNN have access to soil data, while the pretrained transformer models and OLS do not. All models except the OLS use identical temporal windows and train-test splits. 

\section{Experiments}

We test three hypotheses: (1) weather pretraining improves yield prediction on extreme weather years, (2) variational objectives with sinusoidal priors outperform standard approaches, and (3) these benefits generalize to standard years and forward temporal gaps. We evaluate on county-level corn and soybean yield prediction across 763 US Corn Belt counties.

\begin{figure}[htb]
    \centering
    \begin{subfigure}{0.37\textwidth}
        \centering
        \includegraphics[width=\textwidth]{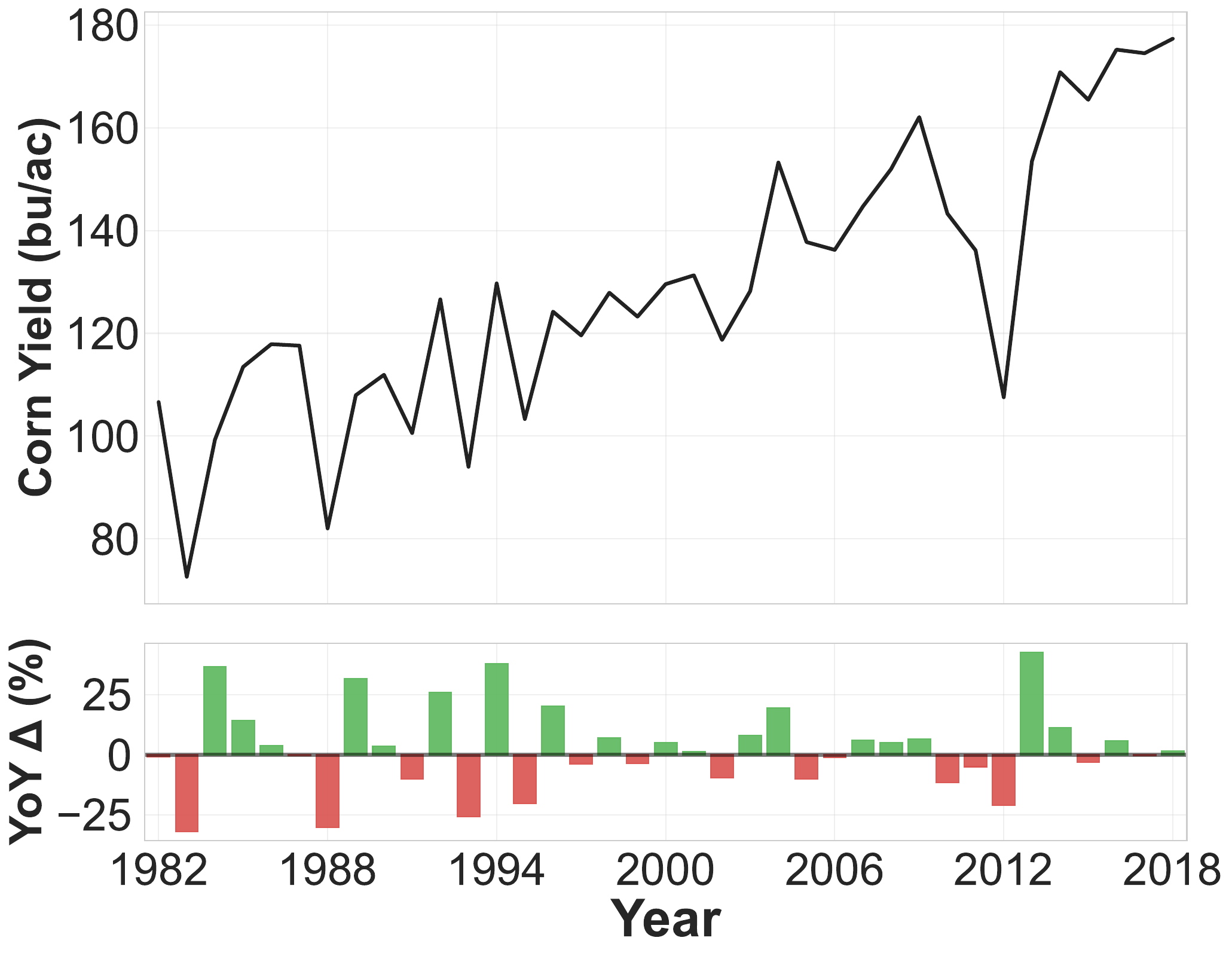}
        \caption{Corn yield trends by year}
        \label{fig:corn_yield_trends}
    \end{subfigure}
    \begin{subfigure}{0.37\textwidth}
        \centering
        \includegraphics[width=\textwidth]{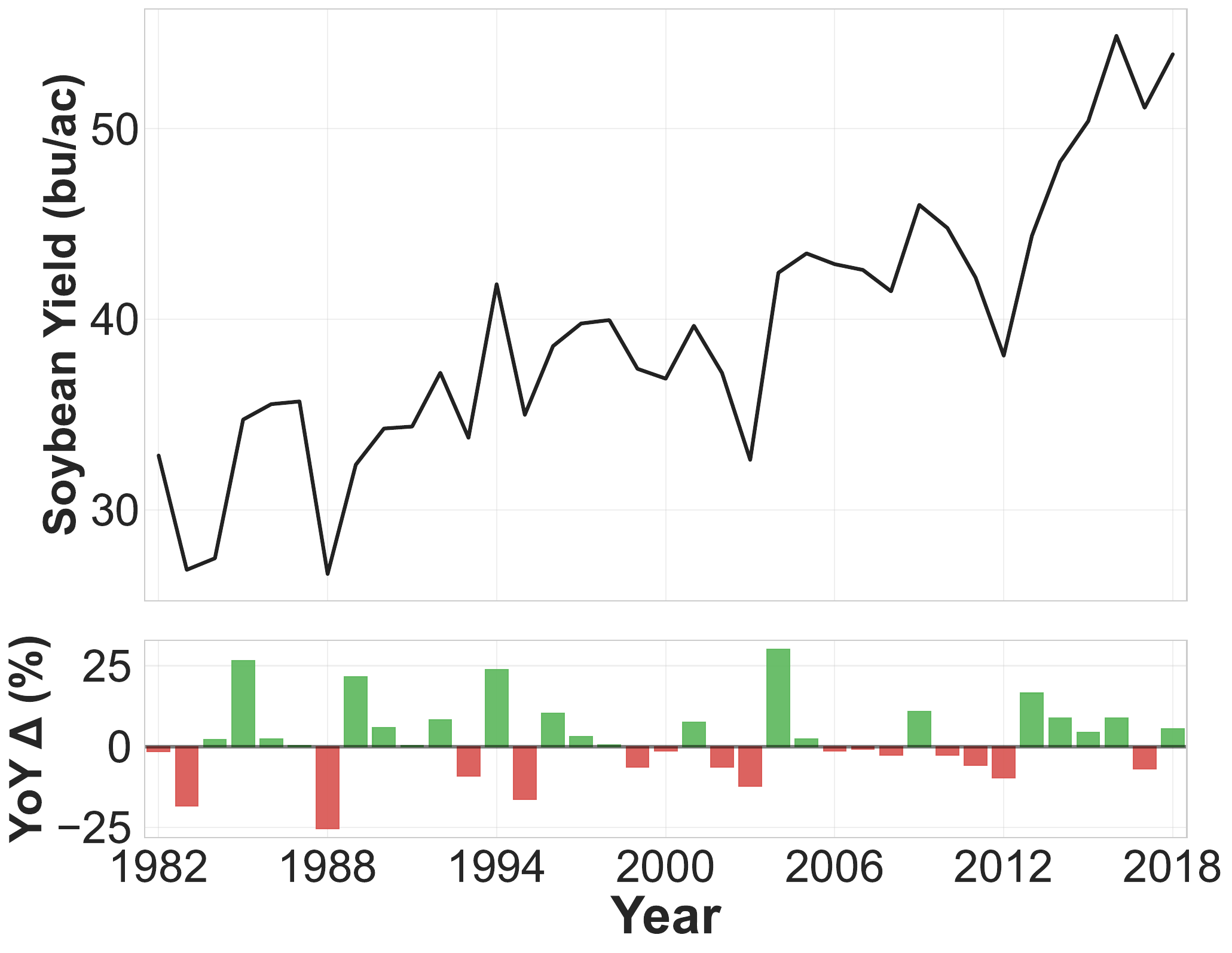}
        \caption{Soybean yield trends by year}
        \label{fig:soybean_yield_trends}
    \end{subfigure}
    \caption{Mean crop yield in bushels per acre (bu/ac) 763 US Corn Belt counties showing extreme weather years as sharp deviations from historical patterns.}
    \label{fig:yield_trends}
\end{figure}

\paragraph{Data.}
We pretrain on the NASA POWER dataset \citep{NASAPower}, comprising 39 years (1984–2022) of climate data at 0.5° resolution across 116 grids over the Americas. It includes 31 meteorological variables aggregated weekly (100K sequences). \textbf{However, its resolution is too coarse for county-level yield forecasting, as the median U.S. county spans 622 square miles} \citep{USCensus_Gazetteer_Counties_2025}. We therefore evaluate on the dataset of \citet{Khaki2019}, containing weather, soil, and county-level corn and soybean yields for 763 Corn Belt counties (1982–2018) \citep{usda_nass_yield}. Corn (C4) and soybean (C3) represent distinct physiological and weather-sensitivity regimes and jointly account for over 60\% of U.S. row-crop acreage \citep{williams2024farmers}.

The \citet{Khaki2019} weather data include six ground-based weekly weather measurements: (1) minimum temperature, (2) maximum temperature, (3) solar radiation, (4) precipitation, (5) snow water equivalent, and (6) vapor pressure, and 11 soil properties averaged over county areas. We exclude soil data to test deployment in data-sparse regions. This presents a key challenge, as pretraining contains satellite measurements of 31 meteorological variables, while fine-tuning uses only 6 ground-based variables averaged over county areas.

\paragraph{Training Configuration.} Full hyperparameters, schedules, and data splits are detailed in Appendix~\ref{sec:training_details}; all models share identical splits and computational budgets. All experiments, including pretraining and grid search, share random seed 1234, unless otherwise noted.

\paragraph{Hyperparameter Optimization.} We performed a 27-configuration grid search to optimize hyperparameters, with full details and robustness analysis in Appendix Figure~\ref{fig:grid_search_supp}. Best hyperparameters are used for all subsequent experiments.

\begin{table*}[htb]
\centering
\begin{tabular}{lccc}
\toprule
\textbf{Method} & \textbf{Corn R² (RMSE)} & \textbf{Soybean R² (RMSE)} & \textbf{Mean R²} \\
\midrule
OLS & 0.227 (27.7) & 0.460 (7.0) & 0.344 \\
XGBoost & 0.135 ± 0.033 (29.0 ± 0.6) & 0.377 ± 0.039 (7.6 ± 0.2) & 0.256 \\
CNN-RNN & 0.256 ± 0.030 (26.5 ± 0.5) & 0.498 ± 0.023 (6.8 ± 0.2) & 0.377 \\
GNN-RNN & 0.564 ± 0.051 (20.2 ± 1.0) & 0.640 ± 0.007 (5.7 ± 0.1) & 0.602 \\
Chronos-Bolt-tiny & 0.525 ± 0.015 (21.6 ± 0.3) & 0.621 ± 0.017 (6.0 ± 0.1) & 0.573 \\
SimMTM & 0.642 ± 0.028 (18.8 ± 0.7) & 0.687 ± 0.018 (5.3 ± 0.1) & 0.665 \\
T-BERT (ours) & 0.660 ± 0.041 (18.3 ± 1.0) & 0.693 ± 0.011 (5.3 ± 0.1) & 0.677 \\
VITA-Std. Normal (ours) & 0.706 ± 0.025 (17.1 ± 0.7) & 0.698 ± 0.020 (5.2 ± 0.2) & 0.702 \\
VITA-Sinusoidal (ours) & \textbf{0.729 ± 0.008 (16.3 ± 0.2)} & \textbf{0.722 ± 0.005 (5.0 ± 0.1)} & \textbf{0.726} \\
\bottomrule
\end{tabular}
\caption{Performance on the 5 most extreme years. Results averaged across 3 random seeds (1234, 5678, 2025) and best results \textbf{bolded}. The OLS baseline shows no change since it is deterministic.}
\label{tab:main_results}
\end{table*}

\subsection{Extreme Year Evaluation (Primary Contribution)}

We identify the five most weather-extreme years for each crop between 2000 and 2018 by computing absolute z-scores from 5-year rolling means of yields. These include known drought years (2002, 2003, 2012) and years with favorable conditions and record-breaking yields (2004, 2009).  \citep{usdm2025}

We also conduct an early-season forecasting experiment using our top four models (two VITA variants, T-BERT, and SimMTM) and OLS, truncating weather data at the end of July in the final year (week 30). This design follows the USDA Economic Research Service (ERS) framework \citep{WestcottJewison2013}, which is conceptually similar to NASS models that also rely on regression-driven estimates grounded in observed weather and crop conditions. The OLS baseline normally incorporates July–August temperature and precipitation for soybean (Appendix~\ref{sec:ols_baseline}), but for consistency across methods we limit it to July in this experiment.

\subsection{Standard Years Generalization}

To validate that extreme weather optimization doesn't compromise standard performance, we evaluate on 2014–2018 using hyperparameters optimized for extreme years. We test both 15-year and 30-year training periods to assess data efficiency requirements.

\subsection{Forward Gap Robustness}

We also evaluate forward gap robustness across five experiments with 5-year gaps: train on 1994–2009/test on 2014, train on 1995–2010/test on 2015, and so forth through 2018.

\subsection{Ablation Studies}

We ablate: (1) pretraining vs. random initialization, (2) variational vs. MSE objective, (3) sinusoidal vs. normal priors, and (4) spatial generalization by pretraining on weather grids excluding continental USA, focusing on extreme weather performance.

All pretraining experiments run on four L40S GPUs and all finetuning experiments run on one L40S GPU with identical computational budgets across methods. The code, pretrained models, and datasets will be publicly released upon publication to facilitate reproducibility and broader agricultural AI research.

\section{Results}
\subsection{Extreme Year Performance}
\begin{figure}[hbt]
\centering
\begin{subfigure}{0.44\textwidth}
    \centering
    \includegraphics[width=\textwidth]{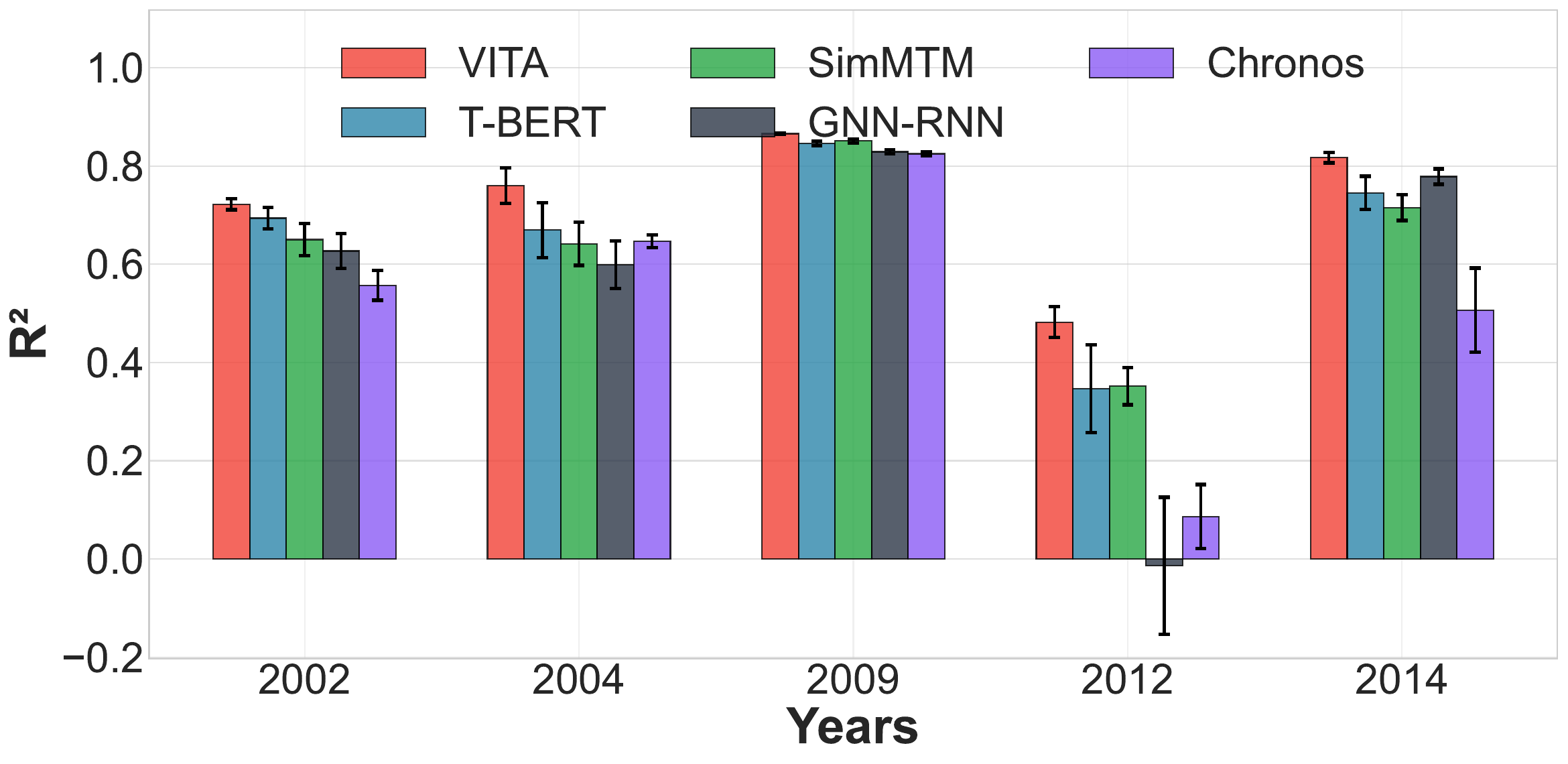}
    \caption{Corn}
    \label{fig:individual_r2_corn_usa}
\end{subfigure}
\\
\begin{subfigure}{0.44\textwidth}
    \centering
    \includegraphics[width=\textwidth]{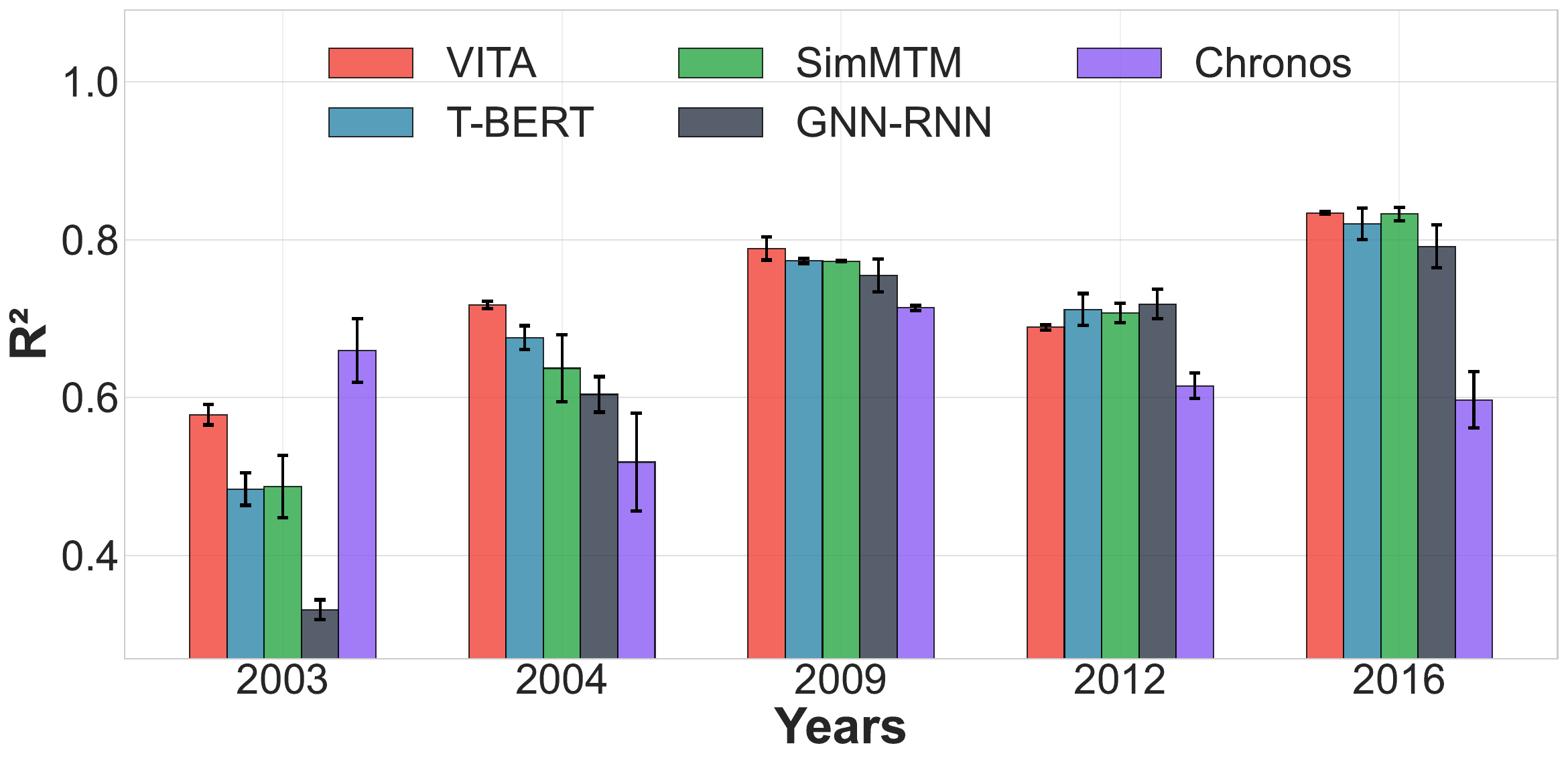}
    \caption{Soybean}
    \label{fig:individual_r2_soybean_usa}
\end{subfigure}
\caption{VITA-Sinusoidal shows consistent improvement over other baselines.}
\label{fig:individual_r2_comparison}
\end{figure}

\begin{table*}[htb]
\centering
\begin{tabular}{lcccc}
\toprule
\textbf{Method} & \textbf{Corn 15yr} & \textbf{Corn 30yr} & \textbf{Soybean 15yr} & \textbf{Soybean 30yr} \\
& R² (RMSE) & R² (RMSE) & R² (RMSE) & R² (RMSE) \\
\midrule
OLS & 0.515 (24.5) & 0.508 (24.1) & 0.673 (6.0) & 0.660 (6.1) \\
XGBoost & 0.439 (27.4) & 0.310 (29.2) & 0.602 (6.5) & 0.564 (7.1) \\
CNN-RNN & 0.659 (20.7) & 0.635 (20.8) & 0.721 (5.6) & 0.671 (6.0) \\
GNN-RNN & 0.788 (16.6) & 0.785 (16.5) & 0.800 (4.7) & 0.810 (4.6) \\
Chronos-Bolt-tiny & 0.704 (19.5) & 0.693 (19.7) & 0.704 (5.7) & 0.724 (5.6) \\
SimMTM & 0.753 (17.9) & 0.768 (17.2) & 0.814 (4.6) & 0.822 (4.5) \\
T-BERT (ours) & 0.791 (16.5) & 0.780 (16.8) & 0.831 (4.4) & \textbf{0.837 (4.3)} \\
VITA-Sinusoidal (ours) & \textbf{0.827 (16.0)} & \textbf{0.837 (15.5)} & \textbf{0.833 (4.3)} & \textbf{0.837 (4.2)} \\
\bottomrule
\end{tabular}
\caption{Standard years (2014-2018) performance. Best results \textbf{bolded}.}
\label{tab:standard_performance}
\end{table*}

VITA-Sinusoidal achieves 0.729 ± 0.008 R² for corn and 0.722 ± 0.005 R² for soybean on the five most extreme weather years (Table~\ref{tab:main_results}), representing +10.5\% and +4.2\% improvements over T-BERT (0.660 and 0.693 respectively). These gains translate to 2.0 bu/ac corn and 0.3 bu/ac soybean RMSE reductions over T-BERT—critically important during droughts like 2012 when accurate forecasts inform billions in crop insurance decisions. The remarkably low variance across random seeds demonstrates the approach's stability.

\begin{table}[htb]
\centering
\begin{tabular}{lccc}
\toprule
\textbf{Crop} & \textbf{Model} & \textbf{Full} & \textbf{Week 30} \\
& & R² (RMSE) & R² (RMSE) \\
\midrule
\multirow{5}{*}{Corn} & OLS & 0.227 (27.7) & 0.227 (27.7) \\
& SimMTM & 0.642 (18.8) & 0.568 (20.5) \\
& T-BERT & 0.660 (18.3) & 0.589 (20.2) \\
& VITA-Std. Norm. & 0.706 (17.1) & 0.642 (18.9) \\
& VITA-Sinusoidal & 0.729 (16.3) & \textbf{0.689 (17.6)} \\
\midrule
\multirow{5}{*}{Soybean} & OLS & 0.460 (7.0) & 0.382 (7.5) \\
& SimMTM & 0.687 (5.3) & 0.481 (6.8) \\
& T-BERT & 0.693 (5.3) & 0.508 (6.7) \\
& VITA-Std. Norm. & 0.698 (5.2) & 0.551 (6.3) \\
& VITA-Sinusoidal & 0.722 (5.0) & \textbf{0.560 (6.2)} \\
\bottomrule
\end{tabular}
\caption{Early-season forecasting on extreme years for the top 4 models and OLS. Full: 7 years weather through end of season (52 weeks). Week 30: cutoff at end of July, ie. week 30 of the final year.}
\label{tab:lead_time}
\end{table}

Figure~\ref{fig:individual_r2_comparison} reveals VITA outperforms T-BERT on 8/10 individual extreme years and surpasses both SimMTM and Chronos-Bolt on 9/10 evaluations (paired t-test across 30 instances from 3 seeds, $p< 0.0001$). The rare underperformances occur when baselines already achieve high accuracy (soybean 2012: 0.689 R², 2016: 0.834 R²), leaving minimal room for improvement. Notably, Chronos-Bolt—despite being 4.5× larger (9M vs 2M parameters) and pretrained 890k sequences \citep{ansari2024chronos}—struggles due to the data asymmetry problem.

We also note that the traditional methods fail catastrophically (OLS: 0.227 corn R², XGBoost: 0.135 R² with soil data) during extreme years, while soil-enriched deep learning shows moderate success (GNN-RNN: 0.564 corn, CNN-RNN: 0.256). Among transformer models, pretraining strategy determines success—SimMTM's temporal masking (0.642 corn) and Chronos's general-purpose approach (0.525 corn) both underperform domain-specific variational pretraining. Comparing T-BERT (0.660) to VITA-Std Normal (0.706) and VITA-Sinusoidal (0.729), we observe that, variational objectives provide +7\% improvement, with sinusoidal priors adding another +3\% by capturing seasonal structure.

Operational viability is confirmed through two stress tests. With weather cut off at the end of July (week 30 of the final year), VITA-Sinusoidal maintains strong performance, achieving 0.689 R² on corn and 0.560 R² -- nearly 3.0× and 1.5× better than OLS baselines, respectively (Table~\ref{tab:lead_time}). The performance drop for soybean is particularly sharp across all models, as its late pod-filling stage makes yields highly sensitive to August rainfall and temperature stress. \cite{WestcottJewison2013}

Lastly, we note that the XGBoost model underperforms in all instances due to correlated weather features, and will require feature engineering to get competitive performance.

Reduced temporal context (5 years vs 7 years) shows VITA-Sinusoidal achieves 0.697 R² corn and 0.684 R² soybean, outperforming most baselines despite using 28\% less data (Appendix~\ref{sec:reduced_context}).

\subsection{Standard Years and Forward Gap Robustness}

Standard years (2014-2018) validate that extreme-weather optimization doesn't compromise performance under normal conditions. VITA achieves 0.827 R² for corn and 0.833 R² for soybean with 15-year training—improvements of +4.6\% and +0.2\% over T-BERT respectively. The contrast with extreme years (+10.5\% corn improvement) supports our core hypothesis: variational uncertainty modeling provides greatest value precisely when predictions are hardest. With 30-year training, corn performance rises to 0.837 R² (+7.3\% over T-BERT), suggesting additional historical data helps but yields diminishing returns compared to better representations.

\begin{table*}[htb]
\centering
\begin{tabular}{lccc}
\toprule
\textbf{Method} & \textbf{Corn} & \textbf{Soybean} & \textbf{Mean R²} \\
& R² (RMSE) & R² (RMSE) & \\
\midrule
OLS & 0.471 (25.6) & 0.634 (6.4) & 0.552 \\
XGBoost & 0.159 (34.4) & 0.433 (8.2) & 0.296 \\
CNN-RNN & 0.556 (23.9) & 0.659 (6.2) & 0.608 \\
GNN-RNN & 0.718 (18.9) & 0.785 (4.9) & 0.752 \\
Chronos-Bolt-tiny & 0.631 (21.7) & 0.685 (5.9) & 0.658 \\
SimMTM & 0.705 (19.7) & 0.776 (5.0) & 0.741 \\
T-BERT (ours) & 0.782 (16.9) & 0.803 (4.7) & 0.793 \\
VITA-Sinusoidal (ours) & \textbf{0.797 (16.3)} & \textbf{0.819 (4.5)} & \textbf{0.808} \\
\bottomrule
\end{tabular}
\caption{Forward gap robustness: 5-year temporal shift (train: 1994-2009, test: 2014-2018). Best results \textbf{bolded}.}
\label{tab:forward_gap}
\end{table*}

Five-year temporal shift (training on 1994-2009, testing on 2014-2018) stresses whether learned patterns generalize beyond training conditions or merely memorize era-specific correlations. VITA achieves 0.797 R² corn and 0.819 R² soybean—maintaining the +1.9\% and +2.0\% margins over T-BERT seen in standard evaluation. Meanwhile, the CNN-RNN and GNN-RNN models suffer noticeable degradation (0.718 corn, down from 0.788), revealing their soil-enhanced features do not help with temporal robustness.

\section{Ablation Studies}
\label{sec:ablation_studies}

\subsection{Pretraining and Spatial Transfer}

\begin{table*}[htb]
\centering
\begin{tabular}{llccc}
\toprule
\textbf{Prior} & \textbf{Pretraining} & \textbf{Corn R²} & \textbf{Soybean R²} & \textbf{t-stat / $p$-value} \\
\midrule
\multirow{3}{*}{Std. Normal} & None & $0.463 \pm 0.230$ & $0.575 \pm 0.134$ & - \\
& Non-US & $0.632 \pm 0.018$ & $0.674 \pm 0.020$ & 3.82 / $7.5\times 10^{-4}$ \\
& Full & $0.706 \pm 0.015$ & $0.698 \pm 0.017$ & 5.49 / $9.3\times 10^{-6}$ \\
\midrule
\multirow{3}{*}{Sinusoidal} & None & $0.469 \pm 0.227$ & $0.569 \pm 0.139$ & - \\
& Non-US & $0.627 \pm 0.020$ & $0.669 \pm 0.021$ & 3.61 / $1.3\times 10^{-3}$ \\
& Full & $0.703 \pm 0.015$ & $0.698 \pm 0.019$ & 5.35 / $1.3\times 10^{-5}$ \\
\bottomrule
\end{tabular}
\caption{VITA pretraining ablation across 27 hyperparameter configurations (random seed 1234).}
\label{tab:pretraining_ablation}
\end{table*}

Pretraining is critical for VITA's performance. Without pretraining, both VITA variants show high variance across hyperparameters (±0.23 corn R², ±0.14 soybean R²) and achieve only ~0.47 and ~0.57 mean R² respectively (Table~\ref{tab:pretraining_ablation}). Pretraining on Central and South American weather—excluding all US continental data—provides substantial improvements of +34-37\% corn and +17-19\% soybean ($t=3.61-3.82, p<0.01$), demonstrating that VITA learns universal weather-agriculture relationships rather than region-specific patterns.

Full Americas pretraining (including US weather outside Corn Belt counties and target years) further improves performance to approximately 0.70 R² for both crops with continued low variance (±0.015-0.019). The +50\% corn and +22\% soybean improvements over no pretraining ($t=5.35-5.49, p<0.001$) highlight that variational objectives are particularly initialization-sensitive—they either converge to strong solutions with pretrained weights or fail to escape poor local minima without them. For comparison, T-BERT shows smaller but still significant pretraining gains (+10.8\% corn, +5.0\% soybean, $t=9.13, p<0.001$) with consistently lower variance across all conditions, as detailed in Appendix~\ref{sec:pretraining_details}.

\section{Discussion}

VITA-Sinusoidal achieves statistically significant improvements over architecturally identical T-BERT baselines ($p < 0.0001$), with gains most pronounced during extreme weather when accurate predictions are most critical. This superior performance stems from rich latent representations that avoid the collapse seen in T-BERT (15.7\% vs. 84.0\% variance in top two PCA components; Figure~\ref{fig:latent_analysis}), enabling better differentiation between normal and extreme conditions. VITA exceeds GNN-RNN despite lacking soil data, demonstrating that historical yields can proxy soil characteristics when paired with rich weather representations.

\paragraph{Operational Context and Social Impact.} Current USDA operational forecasts—including ERS regression models \citep{WestcottJewison2013}—rely on simpler models with hand-crafted features, achieving 0.227 R² for extreme corn years with our OLS baseline. VITA's 3.2× improvement (0.729 R²) translates to substantial impact: 11.4 bu/ac RMSE reduction over OLS and 2.0 bu/ac over T-BERT across 88.7M Corn Belt acres \cite{ncga2024corn}. At \$4.70/bushel \cite{dtn2025insurance}, this translates to \$4.75B and \$800M in value, respectively. This accuracy is critical for the Federal Crop Insurance Program managing billions in premiums \cite{ers2024cropinsurance} and for policy responses during droughts. Deployment requires minimal infrastructure (single GPU, 2.5 hours training) with only public data, enabling integration into existing agricultural statistics systems.

\paragraph{Spatial Transfer and Global Food Security.} Pretraining on Central/South American weather—climatically distinct from the U.S. Corn Belt—significantly improves U.S. predictions ($t=3.61$, $p<0.01$; Table~\ref{tab:pretraining_ablation}). This cross-continental transfer demonstrates that VITA learns universal weather-agriculture relationships (temperature stress, precipitation deficits, radiation anomalies) rather than region-specific patterns. For global food security, models pretrained on data-rich regions can enhance predictions in data-scarce areas despite climatic differences.

\paragraph{Broader Applicability and Limitations.} The decoder-free variational framework can ba applied beyond agriculture to any setting with rich sensors at training and sparse sensors at inference (e.g., ICU vitals vs. labs, IoT vs. industrial telemetry). Current evaluation focuses on U.S. Corn Belt corn and soybean \citep{Khaki2019,gnnrnn2021}; extending to other crops and regions remains future work. We open-sourced implementation, preprocessing scripts, model weights, and documentation for new region adaptation. Data privacy concerns are negligible as we use aggregated public meteorological data, ensuring equitable access through transparency.

\section{Conclusion}

We introduced VITA, a variational pretraining framework for forecasting crop yield under extreme weather. It is pretrained on satellite weather datasets through decoder-free variational learning and achieves state-of-the-art performance on extreme weather years (R² = 0.729 for corn, 0.722 for soybeans). VITA requires only basic weather variables available globally, allowing deployment in data-scarce regions where multi-modal approaches are infeasible. Our comprehensive evaluation across 763 U.S. Corn Belt counties with statistical validation and spatial transfer experiments demonstrates robust, real performance. 

\makeatletter
\@ifundefined{isChecklistMainFile}{
  \newif\ifreproStandalone
  \reproStandalonetrue
}{
  \newif\ifreproStandalone
  \reproStandalonefalse
}
\makeatother

\ifreproStandalone
\documentclass[letterpaper]{article}
\usepackage[submission]{aaai2026}
\setlength{\pdfpagewidth}{8.5in}
\setlength{\pdfpageheight}{11in}
\usepackage{times}
\usepackage{helvet}
\usepackage{courier}
\usepackage{xcolor}
\frenchspacing

\begin{document}
\fi
\setlength{\leftmargini}{20pt}
\makeatletter\def\@listi{\leftmargin\leftmargini \topsep .5em \parsep .5em \itemsep .5em}
\def\@listii{\leftmargin\leftmarginii \labelwidth\leftmarginii \advance\labelwidth-\labelsep \topsep .4em \parsep .4em \itemsep .4em}
\def\@listiii{\leftmargin\leftmarginiii \labelwidth\leftmarginiii \advance\labelwidth-\labelsep \topsep .4em \parsep .4em \itemsep .4em}\makeatother

\setcounter{secnumdepth}{0}
\renewcommand\thesubsection{\arabic{subsection}}
\renewcommand\labelenumi{\thesubsection.\arabic{enumi}}

\newcounter{checksubsection}
\newcounter{checkitem}[checksubsection]

\newcommand{\checksubsection}[1]{%
  \refstepcounter{checksubsection}%
  \paragraph{\arabic{checksubsection}. #1}%
  \setcounter{checkitem}{0}%
}

\newcommand{\checkitem}{%
  \refstepcounter{checkitem}%
  \item[\arabic{checksubsection}.\arabic{checkitem}.]%
}
\newcommand{\question}[2]{\normalcolor\checkitem #1 #2 \color{blue}}
\newcommand{\ifyespoints}[1]{\makebox[0pt][l]{\hspace{-15pt}\normalcolor #1}}

\section*{Reproducibility Checklist}

\vspace{1em}
\hrule
\vspace{1em}

\textbf{Instructions for Authors:}

This document outlines key aspects for assessing reproducibility. Please provide your input by editing this \texttt{.tex} file directly.

For each question (that applies), replace the ``Type your response here'' text with your answer.

\vspace{1em}
\noindent
\textbf{Example:} If a question appears as
\begin{center}
\noindent
\begin{minipage}{.9\linewidth}
\ttfamily\raggedright
\string\question \{Proofs of all novel claims are included\} \{(yes/partial/no)\} \\
Type your response here
\end{minipage}
\end{center}
you would change it to:
\begin{center}
\noindent
\begin{minipage}{.9\linewidth}
\ttfamily\raggedright
\string\question \{Proofs of all novel claims are included\} \{(yes/partial/no)\} \\
yes
\end{minipage}
\end{center}
Please make sure to:
\begin{itemize}\setlength{\itemsep}{.1em}
\item Replace ONLY the ``Type your response here'' text and nothing else.
\item Use one of the options listed for that question (e.g., \textbf{yes}, \textbf{no}, \textbf{partial}, or \textbf{NA}).
\item \textbf{Not} modify any other part of the \texttt{\string\question} command or any other lines in this document.\\
\end{itemize}

You can \texttt{\string\input} this .tex file right before \texttt{\string\end\{document\}} of your main file or compile it as a stand-alone document. Check the instructions on your conference's website to see if you will be asked to provide this checklist with your paper or separately.

\vspace{1em}
\hrule
\vspace{1em}


\checksubsection{General Paper Structure}
\begin{itemize}

\question{Includes a conceptual outline and/or pseudocode description of AI methods introduced}{(yes/partial/no/NA)}
Yes

\question{Clearly delineates statements that are opinions, hypothesis, and speculation from objective facts and results}{(yes/no)}
Yes

\question{Provides well-marked pedagogical references for less-familiar readers to gain background necessary to replicate the paper}{(yes/no)}
Yes

\end{itemize}
\checksubsection{Theoretical Contributions}
\begin{itemize}

\question{Does this paper make theoretical contributions?}{(yes/no)}
Yes

	\ifyespoints{\vspace{1.2em}If yes, please address the following points:}
        \begin{itemize}
	
	\question{All assumptions and restrictions are stated clearly and formally}{(yes/partial/no)}
	Yes

	\question{All novel claims are stated formally (e.g., in theorem statements)}{(yes/partial/no)}
	Yes

	\question{Proofs of all novel claims are included}{(yes/partial/no)}
	Yes

	\question{Proof sketches or intuitions are given for complex and/or novel results}{(yes/partial/no)}
	Yes

	\question{Appropriate citations to theoretical tools used are given}{(yes/partial/no)}
	Yes

	\question{All theoretical claims are demonstrated empirically to hold}{(yes/partial/no/NA)}
	Yes

	\question{All experimental code used to eliminate or disprove claims is included}{(yes/no/NA)}
	NA
	
	\end{itemize}
\end{itemize}

\checksubsection{Dataset Usage}
\begin{itemize}

\question{Does this paper rely on one or more datasets?}{(yes/no)}
Yes

\ifyespoints{If yes, please address the following points:}
\begin{itemize}

	\question{A motivation is given for why the experiments are conducted on the selected datasets}{(yes/partial/no/NA)}
	Yes

	\question{All novel datasets introduced in this paper are included in a data appendix}{(yes/partial/no/NA)}
	Yes

	\question{All novel datasets introduced in this paper will be made publicly available upon publication of the paper with a license that allows free usage for research purposes}{(yes/partial/no/NA)}
	Yes

	\question{All datasets drawn from the existing literature (potentially including authors' own previously published work) are accompanied by appropriate citations}{(yes/no/NA)}
	Yes

	\question{All datasets drawn from the existing literature (potentially including authors' own previously published work) are publicly available}{(yes/partial/no/NA)}
	Yes

	\question{All datasets that are not publicly available are described in detail, with explanation why publicly available alternatives are not scientifically satisficing}{(yes/partial/no/NA)}
	NA

\end{itemize}
\end{itemize}

\checksubsection{Computational Experiments}
\begin{itemize}

\question{Does this paper include computational experiments?}{(yes/no)}
Yes

\ifyespoints{If yes, please address the following points:}
\begin{itemize}

	\question{This paper states the number and range of values tried per (hyper-) parameter during development of the paper, along with the criterion used for selecting the final parameter setting}{(yes/partial/no/NA)}
	Yes

	\question{Any code required for pre-processing data is included in the appendix}{(yes/partial/no)}
	Yes

	\question{All source code required for conducting and analyzing the experiments is included in a code appendix}{(yes/partial/no)}
	Yes

	\question{All source code required for conducting and analyzing the experiments will be made publicly available upon publication of the paper with a license that allows free usage for research purposes}{(yes/partial/no)}
	Yes
        
	\question{All source code implementing new methods have comments detailing the implementation, with references to the paper where each step comes from}{(yes/partial/no)}
	Yes

	\question{If an algorithm depends on randomness, then the method used for setting seeds is described in a way sufficient to allow replication of results}{(yes/partial/no/NA)}
	Yes

	\question{This paper specifies the computing infrastructure used for running experiments (hardware and software), including GPU/CPU models; amount of memory; operating system; names and versions of relevant software libraries and frameworks}{(yes/partial/no)}
	Yes

	\question{This paper formally describes evaluation metrics used and explains the motivation for choosing these metrics}{(yes/partial/no)}
	Yes

	\question{This paper states the number of algorithm runs used to compute each reported result}{(yes/no)}
	Yes

	\question{Analysis of experiments goes beyond single-dimensional summaries of performance (e.g., average; median) to include measures of variation, confidence, or other distributional information}{(yes/no)}
	Yes

	\question{The significance of any improvement or decrease in performance is judged using appropriate statistical tests (e.g., Wilcoxon signed-rank)}{(yes/partial/no)}
	Yes

	\question{This paper lists all final (hyper-)parameters used for each model/algorithm in the paper’s experiments}{(yes/partial/no/NA)}
	Yes

\end{itemize}
\end{itemize}
\ifreproStandalone
\end{document}
\fi

\section*{Appendix}

\subsection{Mathematical Framework}
\label{sec:math_framework}

We define $z \in \mathbb{R}^{364 \times 31}$ as the full set of meteorological variables available during pretraining—such as radiation fluxes, humidity, wind speed, and surface pressure—which together describe the latent physical state of the atmosphere over 364-week sequences (7 years of weekly means). The observed variables $x \in \mathbb{R}^{364 \times 6}$ are a subset of basic weather statistics used in downstream tasks, including quantities like precipitation, and min/max temperature. For brevity, in the following mathematical derivations, we use $d_z = 364 \times 31$ and $d_x = 364 \times 6$ to represent the flattened dimensions. We have two datasets: (1) A large pretraining weather dataset $\mathcal{D}_w = \{(x_i, z_i)\}_{i=1}^{N_w}$ where both basic weather variables $x_i$ and detailed weather states $z_i$ are observed, and (2) A crop yield dataset $\mathcal{D}_y = \{(x_j, y_j)\}_{j=1}^{N_y}$ where only basic weather variables $x_j$ and yield targets $y_j$ are observed, but detailed weather states $z_j$ remain unobserved. Our goal is to predict crop yield $y_j$ given basic weather variables $x_j$ by learning the detailed weather representation $z_j$ during the pretraining phase. Figure~\ref{fig:graphical_model} shows a graphical model of the setup.

A key insight of our framework is the structured relationship between the detailed meteorological state $z$ (e.g., radiation fluxes, humidity) and the basic weather statistics $x$ (e.g., temperature, vapor pressure) used for the downstream task. Many variables in $x$ can be closely approximated or deterministically computed from components in $z$ via our deterministic-decoder assumption.

Inspired by this, we adopt the structural modeling assumption that $p_\theta(x \mid z) \approx 1$ for valid pairs. While not strictly true for all possible weather variables, this is a meteorologically-grounded modeling assumption that is reasonable for our dataset. This assumption allows us to eliminate the decoder term in the variational objective.

\subsubsection{Pretraining Objective}

During pretraining, paired observations $(x_i, z_i) \in \mathcal{D}_w$ can be directly used to fit the likelihood function of $q_\phi(z_i|x_i)$. However, naively fitting likelihood leads to posterior collapse and an overconfident encoder. We therefore employ $\beta$-VAE style objective \citep{higgins2017betavae} that balances reconstruction fidelity with regularization:
\begin{align}
\mathcal{L}_{\text{pretrain}} &= -\mathbb{E}_{(x_i, z_i) \sim \mathcal{D}_w}\Big[\log q_\phi(z_i \mid x_i) \notag \\
&\quad + \alpha \, \text{KL}[q_\phi(z_i \mid x_i) \| p_\theta(z_i)]\Big] \label{eq:pretrain_loss}
\end{align}
where $\alpha > 0$ controls regularization strength. The first term encourages accurate prediction of observed detailed weather states, while the KL term prevents posterior collapse and trains the prior parameters $\theta$.

\subsubsection{Fine-tuning Objective}

Next we derive the variational lower bound for the crop yield prediction. For each sample $(x_j, y_j) \in \mathcal{D}_y$, the log-likelihood is:
\begin{equation}
\log p_\theta(y_j \mid x_j) = \log \int p_\theta(y_j \mid z_j) p_\theta(z_j \mid x_j) dz_j
\end{equation}

Applying Jensen's inequality with variational distribution $q_\phi(z_j \mid x_j)$:
\begin{align}
\log p_\theta(y_j \mid x_j) &\geq \mathbb{E}_{q_\phi(z_j \mid x_j)}[\log p_\theta(y_j \mid z_j)] \notag \\
&\quad - \text{KL}[q_\phi(z_j \mid x_j) \| p_\theta(z_j \mid x_j)] \label{eq:elbo}
\end{align}

Using Bayes' rule: $p_\theta(z_j \mid x_j) = \frac{p_\theta(x_j \mid z_j) p_\theta(z_j)}{p_\theta(x_j)}$ on the KL-Divergence  and ignoring constants, we derive the standard semi-supervised VAE objective \citep{kingma2014semi}:
\begin{align}
\log p_\theta(y_j \mid x_j) &\geq \mathbb{E}_{q_\phi(z_j \mid x_j)}[\log p_\theta(y_j \mid z_j)] \notag \\
&\quad + \mathbb{E}_{q_\phi(z_j \mid x_j)}[\log p_\theta(x_j \mid z_j)] \notag \\
&\quad - \text{KL}[q_\phi(z_j \mid x_j) \| p_\theta(z_j)] \label{eq:full_elbo}
\end{align}

Since by assumption, the detailed weather state $z_j$ deterministically generates basic weather statistics $x_j$, we have $p_\theta(x_j \mid z_j) = 1$ for valid pairs. This eliminates the decoder term $\mathbb{E}_{q_\phi(z_j \mid x_j)}[\log p_\theta(x_j \mid z_j)]$, yielding:
\begin{align}
\mathcal{L}_{\text{yield}} &= -\mathbb{E}_{q_\phi(z_j \mid x_j)}[\log p_\theta(y_j \mid z_j)] \notag \\
&\quad + \text{KL}[q_\phi(z_j \mid x_j) \| p_\theta(z_j)] \label{eq:final_loss}
\end{align}
This loss is computed by first sampling $z_j\sim q_\phi(z_j|x_j)$ with the reparameterization trick \citep{kingma2013auto} and passed to the yield head for yield prediction.

\subsubsection{Distributional Assumptions and Prior Choices}

We assume the variational posterior $q_\phi(z \mid x)$ to be a diagonal Gaussian whose parameters are predicted by a transformer encoder:
\begin{align}
q_\phi(z \mid x) &= \mathcal{N}(z; \mu_\phi(x), \text{diag}(\sigma_\phi^2(x)))
\end{align}

The choice of prior $p_\theta(z)$ depends on the task structure. While standard VAEs \citep{kingma2013auto,kingma2014semi} use unit normal priors, weather data exhibits temporal patterns that motivate more sophisticated choices.

\paragraph{Standard Gaussian Prior.} The simplest choice is $p_\theta(z) = \mathcal{N}(0, I)$.

\paragraph{Sinusoidal Prior.} To capture seasonal patterns:
\begin{align}
p_\theta(z) &= \mathcal{N}(\mu_{\text{sin}}, \text{diag}(\sigma_{\text{sin}}^2)) \\
\mu_{\text{sin},k} &= A_k \sin(\theta_k \cdot \text{pos} + \theta_{0,k})
\end{align}
where parameters $A_k$, $\theta_k$, $\theta_{0,k}$, and $\sigma_{\text{sin},k}^2$ are learned during pretraining for each dimension $k$.

\paragraph{Gaussian Mixture Prior.} While not explored in this paper, a sophisticated prior to capture more complex temporal weather patterns for other fine-tuning tasks could be:
\begin{equation}
p_\theta(z) = \sum_{j=1}^K w_j \mathcal{N}(\mu_{j,\text{sin}}, \text{diag}(\sigma_{j,\text{sin}}^2))
\end{equation}
In this case, the KL term in the loss has no closed form formula and will require Monte Carlo sampling via the reparameterization trick.

\paragraph{Yield Distribution.} Crop yield depends not only on the current year's weather representation $z_j$ but also on historical yield patterns $y_{\text{past},j}$ that capture local soil conditions, management practices, and cultivar effects. For continuous yield targets $y_j$, we assume:
\begin{align}
p_\theta(y_j \mid z_j) &= \mathcal{N}\big( y_j; \mu_{\theta}(z_j, y_{\text{past},j}), \sigma_y^2 \big) \label{eq:yield_likelihood}
\end{align}
where $\mu_{\theta}(z_j, y_{\text{past},j})$ is the mean yield predicted by a neural network with parameters $\theta$, and $\sigma_y^2$ is a noise hyperparameter.

\begin{figure*}[htb]
\centering
\begin{subfigure}{0.32\textwidth}
    \centering
    \includegraphics[width=\textwidth]{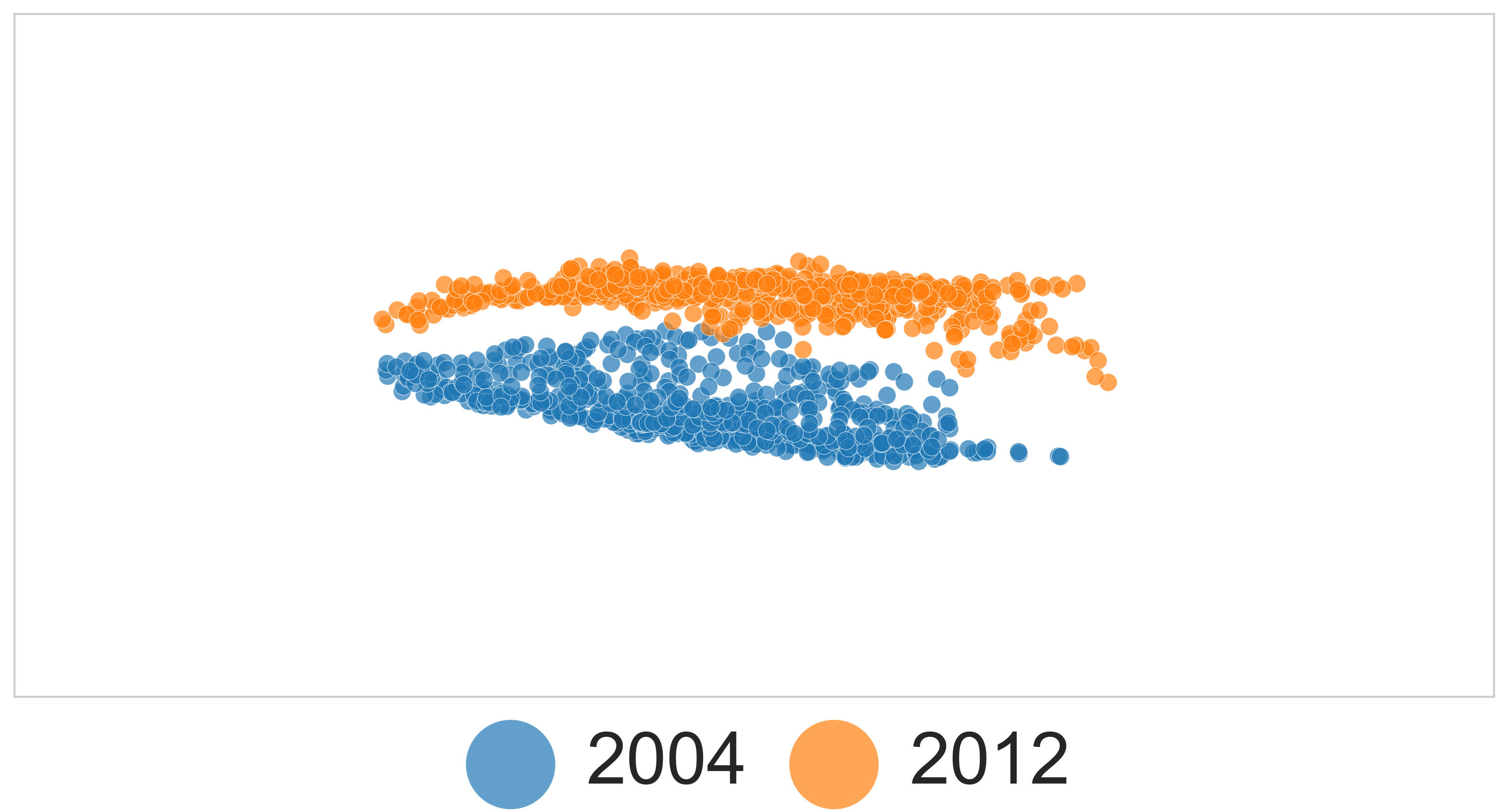}
    \caption{T-BERT (84.0\%)}
    \label{fig:latent_T-BERT}
\end{subfigure}
\begin{subfigure}{0.32\textwidth}
    \centering
    \includegraphics[width=\textwidth]{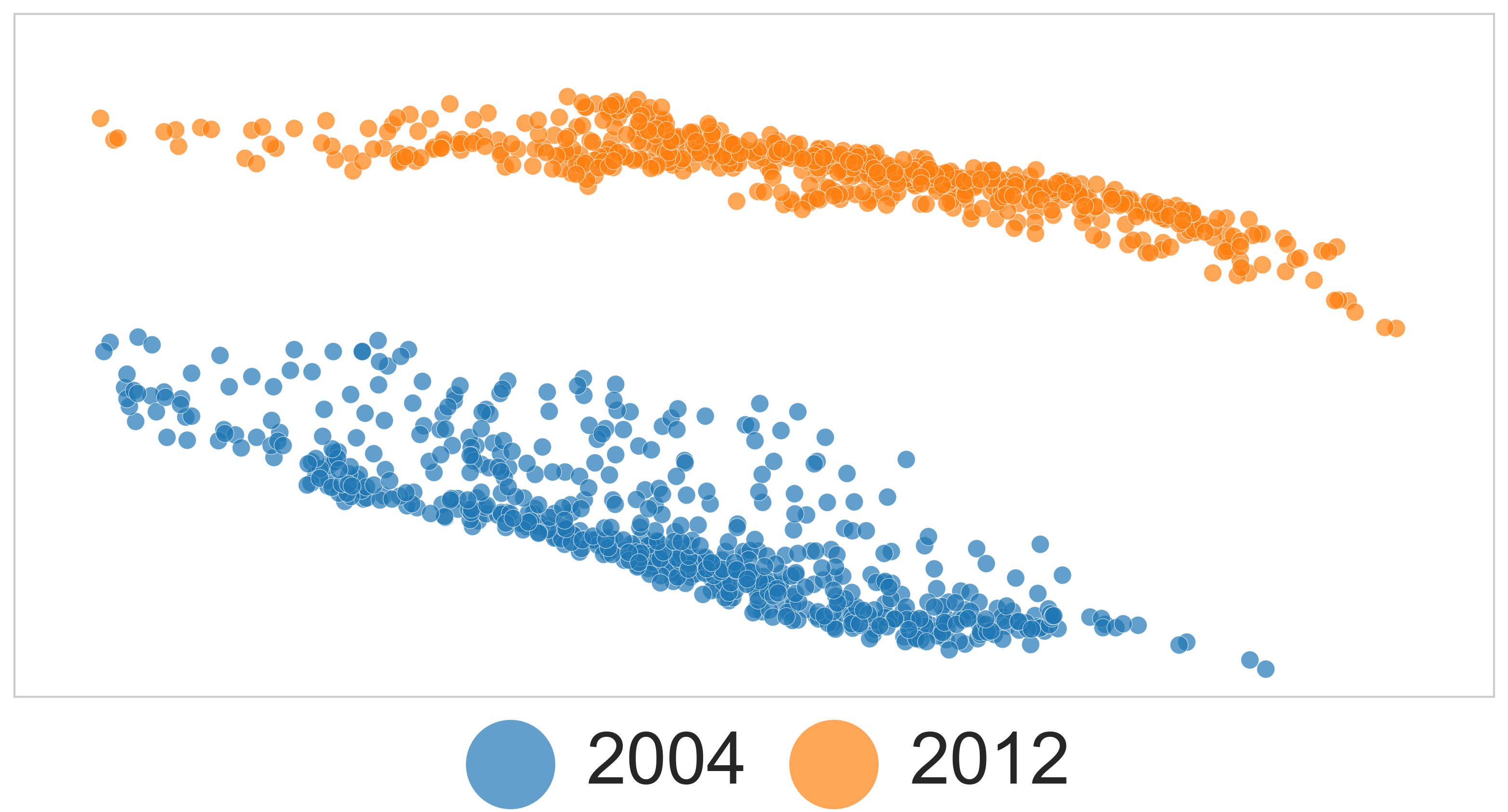}
    \caption{Standard normal prior (37.7\%)}
    \label{fig:latent_normal}
\end{subfigure}
\begin{subfigure}{0.32\textwidth}
    \centering
    \includegraphics[width=\textwidth]{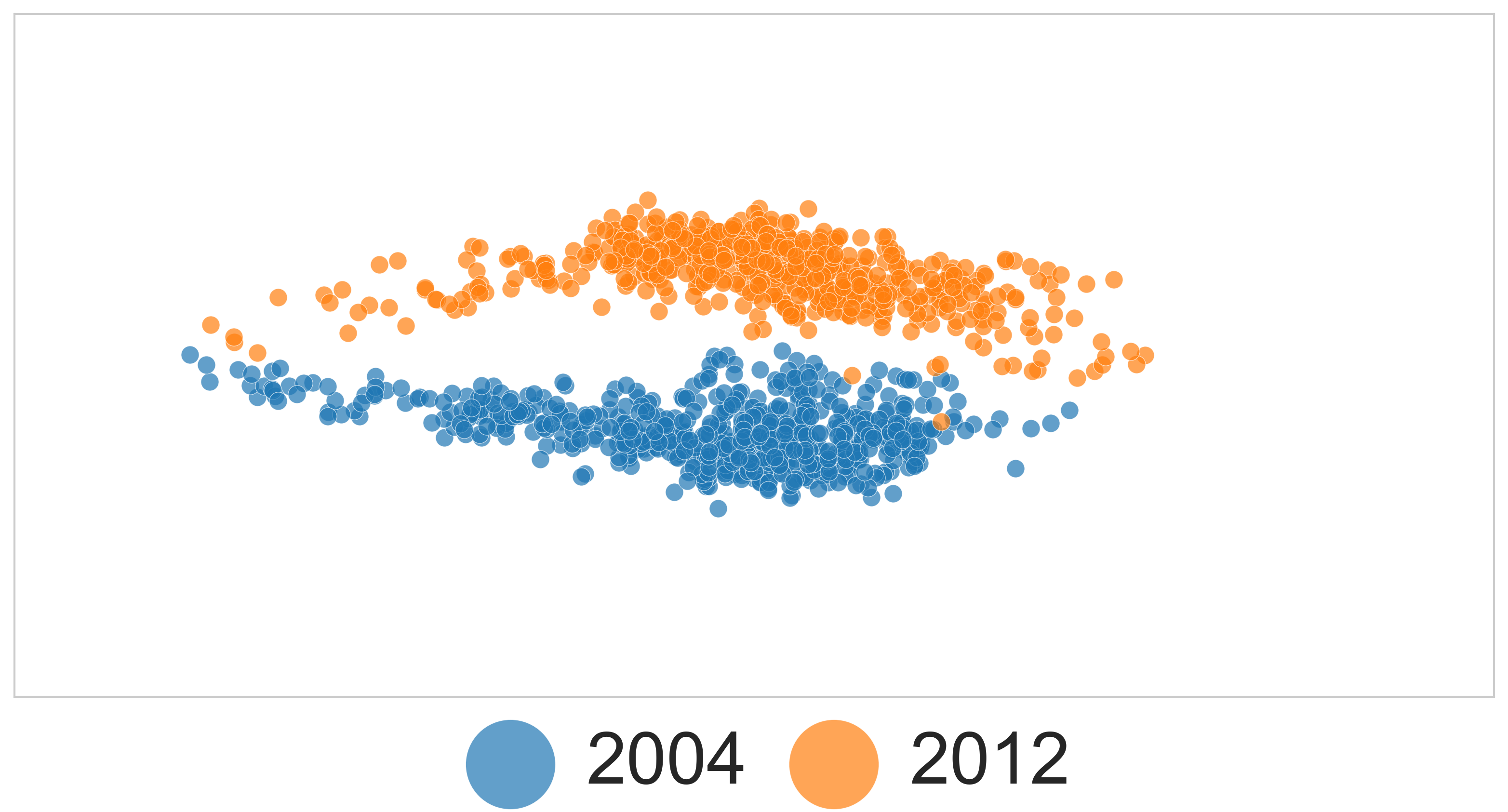}
    \caption{Sinusoidal prior (15.7\%)}
    \label{fig:latent_sinusoidal}
\end{subfigure}
\caption{PCA visualization of the latent weather representations for two extreme years (2004: record-breaking yield, 2012: extreme drought) under different modeling choices. (a) T-BERT (non-variational) shows limited separation and explains 84.0\% of variance in 2D, reflecting a narrow, collapsed latent space. (b) VITA with a standard normal prior yields more separated clusters and explains 34.7\% of variance. (c) Sinusoidal prior induces visually tighter clusters but with only 15.7\% explained variance, indicating even spread of variance into higher-order components.}
\label{fig:latent_analysis}
\end{figure*}

\subsubsection{Practical Fine-tuning Objective}

Substituting the yield likelihood (Equation~(\ref{eq:yield_likelihood})) into the variational lower bound (Equation~(\ref{eq:final_loss})):
\begin{align}
\mathcal{L}_{\text{yield}} &= -\mathbb{E}_{q_\phi(z_j \mid x_j)}[\log p_\theta(y_j \mid z_j)] \notag \\
&\quad + \text{KL}[q_\phi(z_j \mid x_j) \| p_\theta(z_j)] \notag \\
&= \mathbb{E}_{q_\phi(z_j \mid x_j)}\Big[\frac{1}{2\sigma_y^2}(y_j - \mu_\theta(z_j, y_{\text{past},j}))^2\Big] \notag \\
&\quad + \text{KL}[q_\phi(z_j \mid x_j) \| p_\theta(z_j)] + C \label{eq:yield_loss}
\end{align}
where $C$ is a constant. Dropping constants and absorbing $\frac{1}{2\sigma_y^2}$ into the hyperparameter $\beta$:
\begin{align}
\mathcal{L}_{\text{yield}} &= \|y_j - \hat{y_j}\|^2 + \beta \cdot \text{KL}[q_\phi(z_j \mid x_j) \| p_\theta(z_j)] \label{eq:practical_yield_loss}
\end{align}
where $\hat{y_j} = \mathbb{E}_{q_\phi(z_j \mid x_j)}[\mu_{\theta}(z_j, y_{\text{past},j})]$ is the predicted yield. Note that Equation~(\ref{eq:practical_yield_loss}) is still the variational lower bound and the $\beta$ multiplier does not originate from the weakening of the variational objective.

\subsubsection{KL Divergence Computation}

For practical implementation, we need to compute the KL divergence term $\text{KL}[q_\phi(z_j \mid x_j) \| p_\theta(z_j)]$ in both the pretraining and fine-tuning objectives. If both distributions are diagonal Gaussians, this has a closed form.

\paragraph{Standard Gaussian Prior.} When $p_\theta(z) = \mathcal{N}(0, I)$ and $q_\phi(z \mid x) = \mathcal{N}(\mu_\phi(x), \text{diag}(\sigma_\phi^2(x)))$:
\begin{align}
\text{KL}[q_\phi(z \mid x) \| p_\theta(z)] &= \frac{1}{2}\sum_{k=1}^{d_z}\Big[\sigma_{\phi,k}^2(x) + \mu_{\phi,k}^2(x) - 1 \notag \\
&\quad - \log\sigma_{\phi,k}^2(x)\Big] \label{eq:kl_standard}
\end{align}

\paragraph{Sinusoidal Prior.} When $p_\theta(z) = \mathcal{N}(\mu_{\text{sin}}, \text{diag}(\sigma_{\text{sin}}^2))$ with $\mu_{\text{sin},k} = A_k \sin(\theta_k \cdot \text{pos} + \theta_{0,k})$:
\begin{align}
\text{KL}[q_\phi(z \mid x) \| p_\theta(z)] &= \frac{1}{2}\sum_{k=1}^{d_z}\Big[\frac{\sigma_{\phi,k}^2(x)}{\sigma_{\text{sin},k}^2} \notag \\
&\quad + \frac{(\mu_{\phi,k}(x) - \mu_{\text{sin},k})^2}{\sigma_{\text{sin},k}^2} \notag \\
&\quad - 1 - \log\frac{\sigma_{\phi,k}^2(x)}{\sigma_{\text{sin},k}^2}\Big] \label{eq:kl_sinusoidal}
\end{align}

\paragraph{Gaussian Mixture Prior.} For the mixture prior $p_\theta(z) = \sum_{j=1}^K w_j \mathcal{N}(\mu_{j,\text{sin}}, \text{diag}(\sigma_{j,\text{sin}}^2))$, the KL divergence has no closed form and requires Monte Carlo estimation:
\begin{align}
\text{KL}[q_\phi(z \mid x) \| p_\theta(z)] &\approx \frac{1}{L}\sum_{l=1}^L\Big[\log q_\phi(z^{(l)} \mid x) \notag \\
&\quad - \log p_\theta(z^{(l)})\Big] \label{eq:kl_mixture}
\end{align}
where $z^{(l)} \sim q_\phi(z \mid x)$ are samples obtained via the reparameterization trick.

\subsubsection{Latent Space Analysis}
\label{sec:latent_space}

To empirically validate how different prior choices affect the learned representations, we analyze the latent space structure using Principal Component Analysis (PCA) on two extreme years: 2004 (exceptionally good weather with record-breaking yield) and 2012 (record drought with exceptionally bad yield). Figure~\ref{fig:latent_analysis} visualizes these latent structures.

T-BERT, trained without a variational prior, shows a highly compressed latent space with 84.0\% of total variance captured by just two principal components and minimal separation between years—indicating limited representational flexibility. The standard normal prior introduces posterior diversity and improves year-wise separation, but most variance remains concentrated along the x-axis (25.6\%), suggesting over-regularization that forces the model to rely heavily on one latent direction. In contrast, the sinusoidal prior results in only 15.7\% of variance explained by the top two components (14.0\% and 1.7\%), despite having tighter clustering. This indicates the sinusoidal model distributes information more evenly across the full latent space, yielding richer representations. These findings confirm that seasonality-aware priors enforce meaningful structure without excessive compression.

\subsection{Experimental Setup and Implementation Details}
\label{sec:experimental_setup}

This section provides comprehensive details on datasets, evaluation protocols, model configurations, and computational requirements to ensure full reproducibility.

\subsubsection{Dataset Spatial Coverage}
\label{sec:spatial_coverage}

\begin{figure*}[htbp]
    \centering
    \begin{subfigure}{0.4\textwidth}
        \centering
        \includegraphics[width=\textwidth]{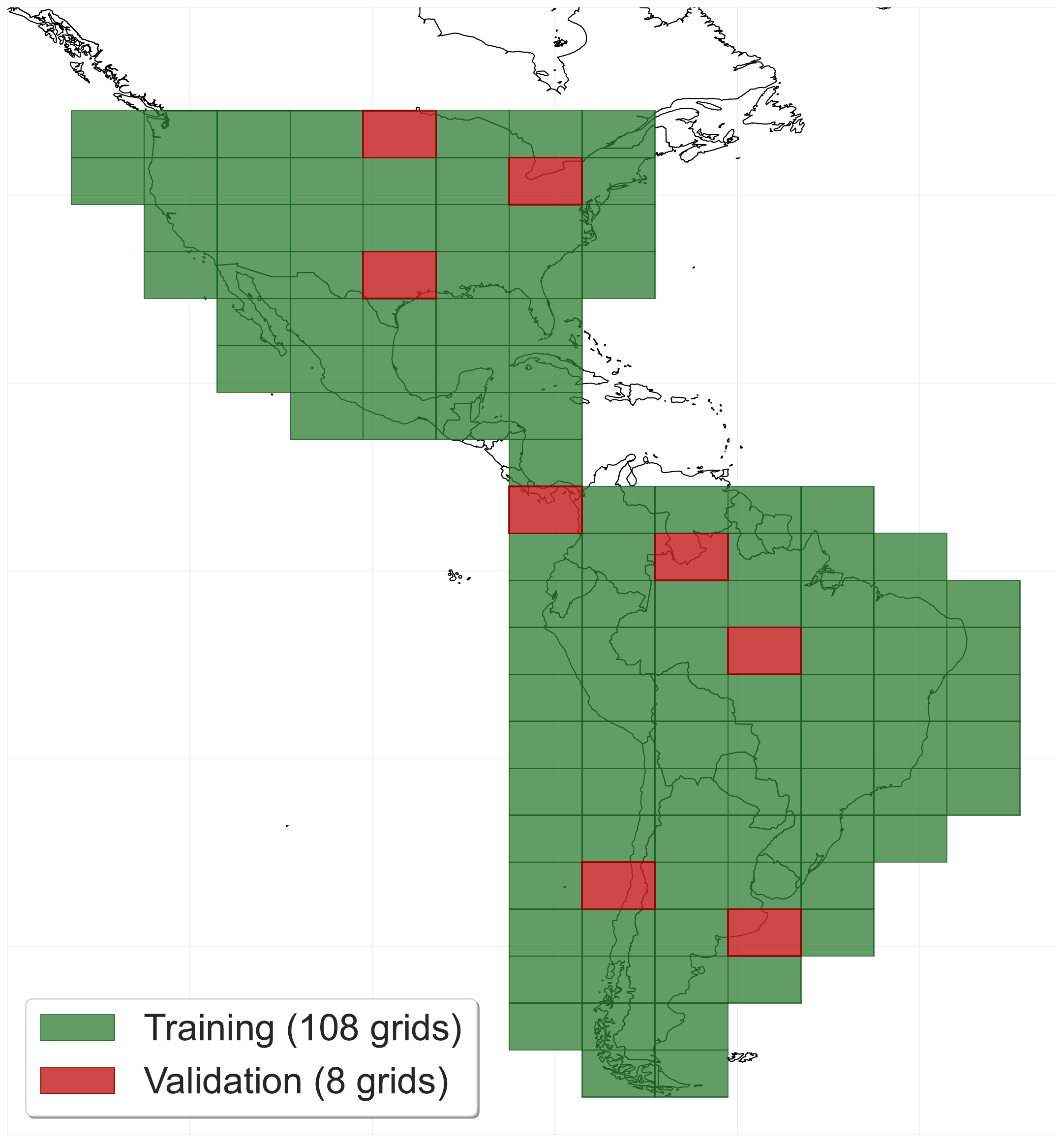}
        \caption{Pretraining weather data coverage.}
        \label{fig:pretraining-weather-map}
    \end{subfigure}
    \begin{subfigure}{0.45\textwidth}
        \centering
        \includegraphics[width=\textwidth]{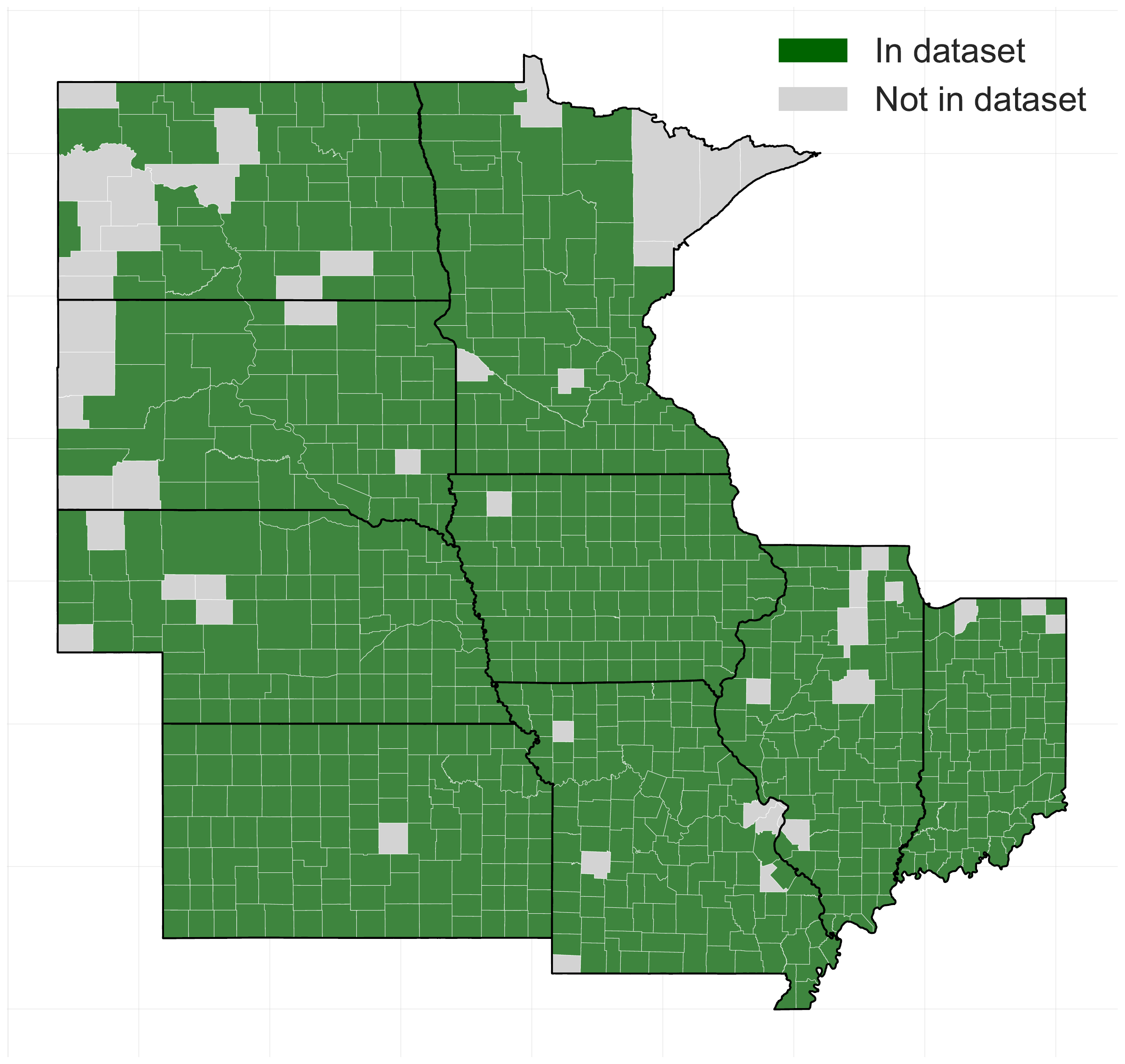}
        \caption{Fine-tuning crop yield data coverage.}
        \label{fig:finetuning-counties-map}
    \end{subfigure}
    \caption{Spatial coverage of pretraining and fine-tuning datasets.}
    \label{fig:dataset-maps}
    \end{figure*}

\paragraph{Locations.} Our approach leverages two distinct spatial datasets with different coverage and resolution. The pretraining dataset contains 116 spatial grids at 0.5° resolution covering the continental United States, Central America and South America, with 108 grids used for training and 8 grids for validation. The fine-tuning dataset focuses on 763 counties within the U.S. Corn Belt region, representing the core agricultural areas for corn and soybean production. Figure~\ref{fig:dataset-maps} illustrates the spatial coverage of both datasets, demonstrating how weather knowledge learned from the broader Americas region transfers effectively to US agricultural counties.

\paragraph{Pretraining Dataset Variables.} Table~\ref{tab:weather_variables} shows a list of 31 weather measurements in the pretraining data. The first 28 measurements were downloaded from the NASA Power Project \citep{NASAPower} from 1984 to 2022. The last three were predicted using Tetens equation (Vapor Pressure and Vapor Pressure Deficit) and FAO-Penman-Monteith equation (Reference Evapotranspiration). The data was downloaded as daily measurements and weekly mean was computed for each variable.

\subsubsection{Extreme Years Selection and Evaluation Protocol}
\label{sec:extreme_years}

To rigorously evaluate model performance on unprecedented weather conditions, we identify extreme years using standardized deviations from historical norms. Table~\ref{tab:extreme_years_selection} presents our selection methodology: years are ranked by absolute z-score from their 5-year rolling mean yield, with the top 5 years per crop selected for evaluation. This captures both exceptionally favorable years (e.g., corn 2004 with +21.42\% yield, z-score 5.25) and severe stress years (e.g., the 2012 drought with -27.15\% corn yield, z-score 4.08).

\begin{table*}[htb]
\centering
\begin{tabular}{lccccc}
\toprule
\textbf{Crop} & \textbf{Year} & \textbf{Yield (Bu/Acre)} & \textbf{5-Year Mean} & \textbf{Deviation \%} & \textbf{Abs. Z-Score} \\
\midrule
\multirow{5}{*}{\textbf{Corn}}
& 2004 & 153.25 & 126.21 & 21.42 & 5.25 \\
& 2012 & 107.55 & 147.64 & 27.15 & 4.08 \\
& 2009 & 162.09 & 144.78 & 11.96 & 2.21 \\
& 2002 & 118.71 & 126.32 & 6.03 & 1.58 \\
& 2014 & 170.83 & 140.52 & 21.58 & 1.45 \\
\midrule
\multirow{5}{*}{\textbf{Soybean}}
& 2009 & 46.01 & 42.57 & 8.07 & 4.72 \\
& 2003 & 32.63 & 38.22 & 14.63 & 3.81 \\
& 2012 & 38.10 & 43.41 & 12.24 & 2.78 \\
& 2004 & 42.44 & 36.75 & 15.48 & 2.23 \\
& 2016 & 54.88 & 44.67 & 22.85 & 2.09 \\
\bottomrule
\end{tabular}
\caption{Top 5 extreme years selection based on yield deviation from 5-year rolling mean.}
\label{tab:extreme_years_selection}
\end{table*}

\paragraph{Strict Hold-Out Protocol.} For each extreme test year $t_{\text{test}}$, we train exclusively on the 15 preceding years $[t_{\text{test}}-15, t_{\text{test}}-1]$ and evaluate only on $t_{\text{test}}$. Training generates overlapping 7-year sequences with target years in $[t_{\text{test}}-9, t_{\text{test}}-1]$, ensuring $t_{\text{test}}$ never appears during training. For example, when evaluating corn 2002, we train on 1987-2001 generating sequences $(1987\text{-}1993) \rightarrow 1993$, ..., $(1995\text{-}2001) \rightarrow 2001$, while $(1996\text{-}2002) \rightarrow 2002$ is only used for final evaluation.

\subsubsection{Training Configuration Details}
\label{sec:training_details}

\paragraph{Pretraining Configuration.} We pretrain on weather sequences using batch size 256, and learning rate $5\times 10^{-4}$. Progressive masking increases from 10 to 25 features over 100 epochs, with 10-epoch linear warmup followed by exponential decay ($\delta=0.99$). The variational objective uses $\alpha = 0.5$ to balance reconstruction and KL regularization.

\paragraph{Fine-tuning Configuration.} Following a similar setup to \citet{Khaki2019} and \citet{gnnrnn2021}, we construct overlapping 7-year sequences $(t-6, ..., t)$ for each county, generating 9 sequences from 15-year training windows. Each sequence predicts yield for year $t$ using weather from all 7 years plus historical yields from $t-6$ to $t-1$. Training proceeds for 40 epochs with 10-epoch linear warmup and cosine annealing, selecting the best validation RMSE. We use 15-year windows for extreme year and robustness experiments, and test both 15-year and 30-year windows for standard years, with strict isolation between test runs to prevent data leakage.

\paragraph{Random Seeds.} All experiments, unless noted otherwise, share random seed 1234. The best extreme year results were additionally validated across 3 seeds (1234, 5678, 2025) and averaged for robustness.

\subsubsection{VITA Architecture Details}
\label{sec:architecture_details}

Table~\ref{tab:architecture} details the VITA architecture. The core transformer encoder uses 4 layers with 10 attention heads, hidden dimension $d_h=200$, and MLP dimension $d_{\text{mlp}}=800$, processing weather sequences. The weather output projection produces mean and variance ($31\times 2$) for the variational posterior. Weather attention aggregates 31 feature dimensions using a learned attention mechanism (hidden dim 16). The sinusoidal prior parameterizes amplitude, frequency, phase, and variance for 31 weather variables across 364 time positions ($124 \times 364$ total parameters). All MLPs use GELU activation \citep{hendrycks2017bridging}.

\begin{table}[htb]
\centering
\begin{tabular}{lccc}
\toprule
\textbf{Component} & \textbf{Input} & \textbf{Hidden} & \textbf{Output} \\
\midrule
Weather Input Proj. & $34$ & - & $200$ \\
Transformer ($\times 4$) & $200$ & $800$ & $200$ \\
Weather Output Proj. & $200$ & $-$ & $31\times 2$ \\
Weather Attention & $31$ & $16$ & $1$ \\
Yield MLP & $36$ & $120$ & $1$ \\
Sine Prior Param. & - & - & $124 \times 364$\\
\bottomrule
\end{tabular}
\caption{VITA Model Architecture}
\label{tab:architecture}
\end{table}

\subsubsection{Computational Efficiency}
\label{sec:runtime_analysis}

Table~\ref{tab:efficiency} presents runtime comparisons demonstrating that VITA's variational objective adds minimal computational overhead. Fine-tuning requires 25.53 minutes per configuration—only 3.4\% slower than non-variational T-BERT (24.69 min) while delivering statistically significant performance gains. SimMTM requires more pretraining time (32.56 min vs 27-29 min) due to its sophisticated temporal masking strategy, whereas VITA and T-BERT employ simpler feature-wise masking.

\begin{table}[htb]
\centering
\begin{tabular}{lcc}
\toprule
\textbf{Method} & \textbf{Pretraining} & \textbf{Fine-tuning} \\
& $4\times$ L40S & $1\times$ L40S\\
\midrule
OLS & N/A & 6.4 ± 1.3 min \\
XGBoost & N/A & 18.7 ± 1.4 min \\
CNN-RNN & N/A & 8.58 ± 0.04 min \\
GNN-RNN & N/A & 7.13 ± 0.03 min \\
Chronos-Bolt-tiny & N/A & 44.7 ± 1.3 min \\
SimMTM & 32.56 min & 25.92 ± 0.01 min \\
T-BERT & 27.42 min & 24.69 ± 0.13 min \\
VITA-Sinusoidal & 29.21 min & 25.53 ± 0.58 min \\
\bottomrule
\end{tabular}
\caption{Runtime comparison on 4×L40S GPUs (pretraining) and 1×L40S GPU (fine-tuning).}
\label{tab:efficiency}
\end{table}

Pretraining on 4×L40S GPUs takes 27-29 minutes, extrapolating to ~108-116 minutes on a single L40S. Combined with 25-minute fine-tuning, the complete pipeline requires ~133-141 minutes ($<2.5$ hours) on a single GPU. The full hyperparameter grid search (27 configurations × 2 crops = 54 experiments) totals ~22.5-23.4 hours of fine-tuning time.

\subsection{Complete Experimental Results and Analysis}
\label{sec:complete_results}

\begin{figure*}[htb]
\centering
\begin{subfigure}{\textwidth}
    \centering
    \includegraphics[width=0.95\textwidth]{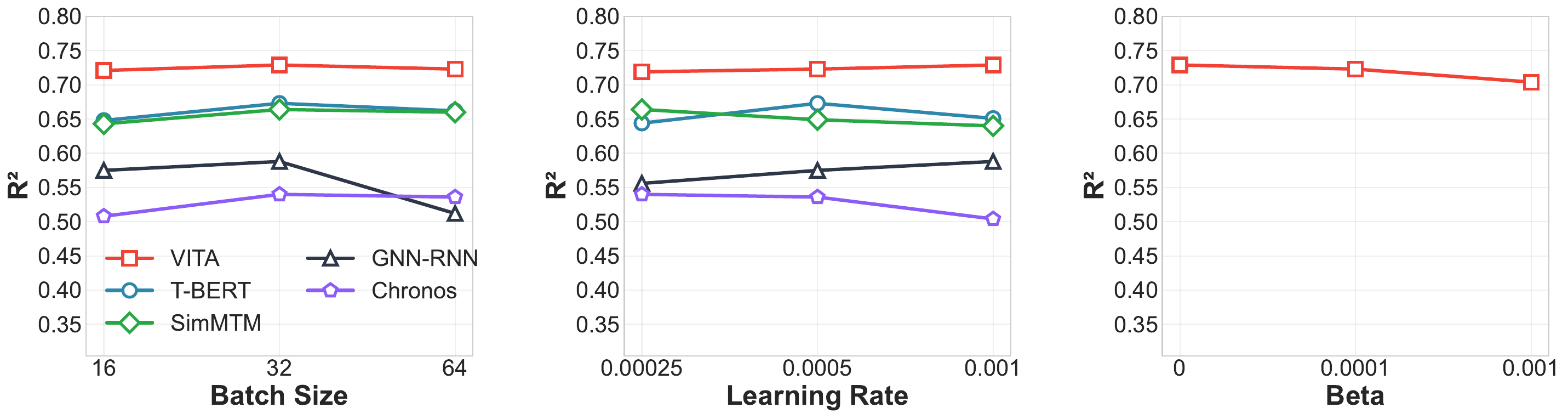}
    \caption{Corn hyperparameter grid search}
    \label{fig:grid_search_corn_supp}
\end{subfigure}
\\[0.5em]
\begin{subfigure}{\textwidth}
    \centering
    \includegraphics[width=0.95\textwidth]{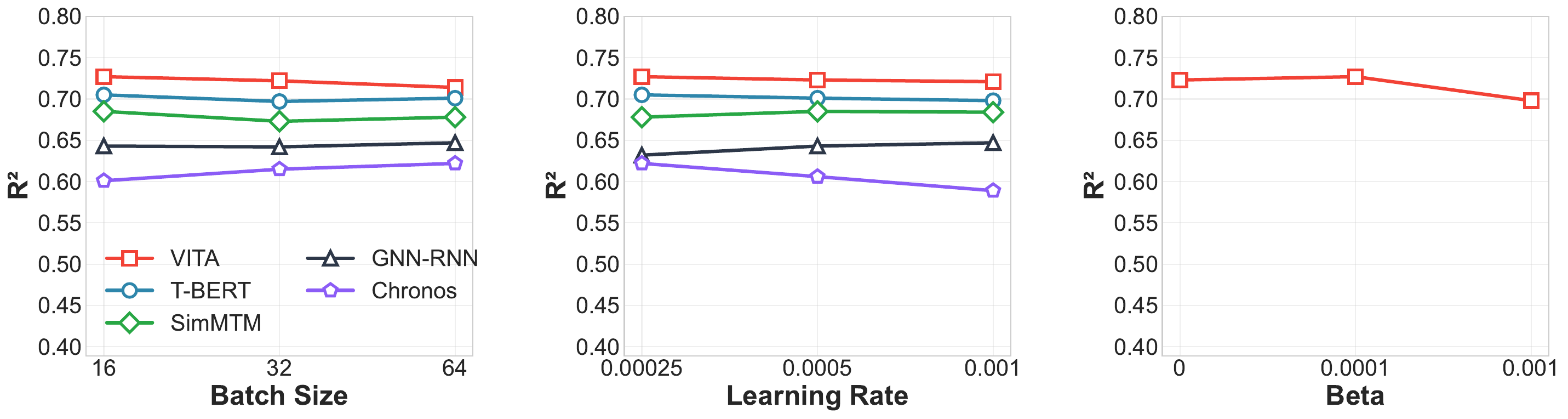}
    \caption{Soybean hyperparameter grid search}
    \label{fig:grid_search_soybean_supp}
\end{subfigure}
\caption{Hyperparameter robustness on extreme years: VITA-Sinusoidal consistently outperforms other baselines across all configurations tested. The CNN-RNN performs considerably worse and is not shown in the plot.}
\label{fig:grid_search_supp}
\end{figure*}

\subsubsection{Impact of Pretraining Across Hyperparameter Configurations}
\label{sec:pretraining_details}

\begin{table*}[htbp]
\centering
\begin{tabular}{llccccc}
\toprule
\textbf{Crop} & \textbf{Model} & \textbf{Pretrained} & \textbf{Best R²} & \textbf{Mean R²} & \textbf{t-statistic} & \textbf{p-value} \\
\midrule
\multirow{6}{*}{Corn} & \multirow{2}{*}{T-BERT} & No & 0.604 & $0.584 \pm 0.015$ & \multirow{2}{*}{9.13} & \multirow{2}{*}{$1.7\times 10^{-5}$} \\
& & Yes & 0.660 & $0.647 \pm 0.014$ & & \\
\cmidrule{2-7}
& \multirow{2}{*}{VITA-Std} & No & 0.667 & $0.463 \pm 0.230$ & \multirow{2}{*}{5.49} & \multirow{2}{*}{$9.3\times 10^{-6}$} \\
& & Yes & 0.706 & $0.706 \pm 0.015$ & & \\
\cmidrule{2-7}
& \multirow{2}{*}{VITA-Sinusoidal} & No & 0.672 & $0.469 \pm 0.227$ & \multirow{2}{*}{5.35} & \multirow{2}{*}{$1.3\times 10^{-5}$} \\
& & Yes & 0.729 & $0.703 \pm 0.015$ & & \\
\midrule
\multirow{6}{*}{Soybean} & \multirow{2}{*}{T-BERT} & No & 0.680 & $0.660 \pm 0.020$ & \multirow{2}{*}{4.36} & \multirow{2}{*}{$2.4\times 10^{-3}$} \\
& & Yes & 0.693 & $0.693 \pm 0.012$ & & \\
\cmidrule{2-7}
& \multirow{2}{*}{VITA-Std} & No & 0.689 & $0.575 \pm 0.134$ & \multirow{2}{*}{4.76} & \multirow{2}{*}{$6.3\times 10^{-5}$} \\
& & Yes & 0.698 & $0.698 \pm 0.017$ & & \\
\cmidrule{2-7}
& \multirow{2}{*}{VITA-Sinusoidal} & No & 0.684 & $0.569 \pm 0.139$ & \multirow{2}{*}{4.80} & \multirow{2}{*}{$5.7\times 10^{-5}$} \\
& & Yes & 0.722 & $0.698 \pm 0.019$ & & \\
\bottomrule
\end{tabular}
\caption{Pretraining effect across hyperparameter configurations during grid search. T-BERT evaluated over 9 configurations (3 learning rates × 3 batch sizes × 1 regularization value), VITA variants over 27 configurations (3 learning rates × 3 batch sizes × 3 $\beta$ values). P-values from paired t-tests comparing pretrained vs non-pretrained.}
\label{tab:pretraining_ablation_detailed}
\end{table*}

Tables~\ref{tab:pretraining_ablation} and \ref{tab:pretraining_ablation_detailed} comprehensively evaluate pretraining benefits across hyperparameter configurations (seed 1234). T-BERT shows consistent but moderate gains: +10.8\% mean corn R² (0.647 vs 0.584, $t=9.13$, $p<0.001$) and +5.0\% soybean R² (0.693 vs 0.660, $t=4.36$, $p<0.01$), with low variance (±0.014-0.020) indicating stable, initialization-insensitive optimization.

In contrast, VITA variants exhibit dramatic improvements. Without pretraining, both VITA-Std. Normal and VITA-Sinusoidal show extremely high variance (±0.227-0.230), indicating unstable optimization. Pretraining reduces variance by over 10× while boosting mean R² by +49-52\% for corn and +21-23\% for soybean (all $p<0.001$), demonstrating that variational objectives critically depend on well-initialized priors.

\paragraph{Cross-Continental Transfer.} Table~\ref{tab:pretraining_ablation} validates spatial generalization by pretraining exclusively on Central/South American weather (excluding all continental US data). Despite zero geographic overlap, pretraining still provides substantial gains: +33-37\% corn R² and +17-19\% soybean R² ($t=3.61-3.82$, $p<0.01$). This confirms VITA learns universal weather-agriculture relationships (e.g., temperature stress effects on photosynthesis) rather than location-specific patterns—critical for deployment in data-scarce regions.

\subsubsection{Hyperparameter Robustness Analysis}

Figure~\ref{fig:grid_search_supp} visualizes performance across 27 deep learning configurations: learning rates $\{2.5 \times 10^{-4},\, 5 \times 10^{-4},\, 1 \times 10^{-3}\}$ × batch sizes $\{16, 32, 64\}$ × regularization $\beta \in \{0, 10^{-4}, 10^{-3}\}$. XGBoost uses an equivalent 27-configuration search over estimators $\{100, 500, 1000\}$ × depth $\{4, 6, 8\}$ × learning rate $\{0.03, 0.05, 0.1\}$.

VITA-Sinusoidal consistently outperforms T-BERT across \emph{all} configurations, with $R^2$ improvements ranging from $+0.01$ to $+0.15$. This systematic advantage across diverse hyperparameter settings demonstrates that performance gains are intrinsic to the variational framework rather than artifacts of hyperparameter tuning.

\subsubsection{Statistical Significance Analysis}
\label{sec:statistical_analysis}

\begin{table}[htb]
\centering
\begin{tabular}{lccc}
\toprule
\textbf{Crop} & \textbf{Year} & \textbf{T-BERT} & \textbf{VITA-Sinusoidal} \\
 & & \textbf{R² ± std} & \textbf{R² ± std} \\
\midrule
\multirow{5}{*}{\textbf{Corn}} 
& 2002 & $0.694 \pm 0.021$ & $0.722 \pm 0.011$ \\
& 2004 & $0.670 \pm 0.056$ & $0.760 \pm 0.036$ \\
& 2009 & $0.846 \pm 0.005$ & $0.866 \pm 0.002$ \\
& 2012 & $0.347 \pm 0.089$ & $0.482 \pm 0.031$ \\
& 2014 & $0.745 \pm 0.034$ & $0.817 \pm 0.010$ \\
\midrule
\multirow{5}{*}{\textbf{Soybean}} 
& 2003 & $0.484 \pm 0.021$ & $0.579 \pm 0.013$ \\
& 2004 & $0.676 \pm 0.015$ & $0.718 \pm 0.005$ \\
& 2009 & $0.773 \pm 0.003$ & $0.789 \pm 0.015$ \\
& 2012 & $0.712 \pm 0.020$ & $0.689 \pm 0.003$ \\
& 2016 & $0.820 \pm 0.020$ & $0.834 \pm 0.002$ \\
\midrule
\textbf{Mean} & & \textbf{0.677} & \textbf{0.726} \\
\bottomrule
\end{tabular}
\caption{Detailed R² performance on extreme years for statistical analysis.}
\label{tab:detailed_r2}
\end{table}

Table~\ref{tab:detailed_r2} presents complete R² scores across extreme years and crops, enabling rigorous statistical evaluation. We conduct both parametric and non-parametric tests on 30 paired comparisons (2 crops × 5 extreme years × 3 random seeds). Paired t-test: $t = 4.91$, $p < 0.0001$. Wilcoxon Signed-Rank Test: $W = 37.0$, $p < 0.0001$. VITA-Sinusoidal outperforms T-BERT in 25 of 30 comparisons, with mean R² improvement from 0.677 to 0.726 (+7.2\% relative). Both tests confirm highly significant improvements.

\subsubsection{Year-Permutation Ablation Study}
\label{sec:year_permutation}

\begin{table}[htb]
\centering
\begin{tabular}{lcccc}
\toprule
\textbf{Crop} & \textbf{Prior} & \textbf{No-Pre.} & \textbf{Pre.} & \\
 & & \textbf{Best R²} & \textbf{Best R²} & \textbf{$\Delta$} \\

\midrule
\multirow{2}{*}{\textbf{Corn}} & Std-Norm. & $0.685$ & $0.732$ & +0.047 \\
& Sinusoidal & $0.672$ & $0.708$ & +0.036 \\
\midrule
\multirow{2}{*}{\textbf{Soybean}} & Std-Norm. & $0.687$ & $0.717$ & +0.030 \\
& Sinusoidal & $0.684$ & $0.715$ & +0.031 \\
\bottomrule
\end{tabular}
\caption{Year-permutation ablation on corn and soybean yield prediction. Pretraining benefits persist even when temporal relationships are destroyed.}
\label{tab:year_permutation}
\end{table}

To rule out year-specific memorization, we conduct an ablation where calendar-year labels in the pretraining dataset (1984-2022) are randomly permuted while keeping weather measurements unchanged. This destroys temporal relationships, forcing the model to learn from weather statistics alone. Table~\ref{tab:year_permutation} shows that pretraining still provides $\geq +0.03$ improvements with fixed hyperparameters, confirming benefits arise from weather representation learning rather than temporal memorization.

Critically, the sinusoidal prior shows reduced gains in the permuted setting (corn: +0.036 vs standard normal +0.047). This occurs because sinusoidal priors impose sine-wave structure to capture seasonality—when temporal order is randomized, this structure becomes counterproductive. This validates two key insights: (1) VITA's pretraining learns robust weather patterns, not temporal correlations, and (2) the sinusoidal prior's advantages require preserved temporal structure, confirming its role as a seasonality-aware regularizer.

\subsection{Additional Baseline Experiments and Robustness Tests}
\label{sec:additional_experiments}

Beyond the comprehensive ablations above, we evaluate VITA against domain-engineered baselines and stress-test performance under operational constraints (incomplete seasonal data, reduced temporal context).

\subsubsection{OLS Baseline: Domain-Engineered Features}
\label{sec:ols_baseline}

To establish a domain-grounded comparison, we implement a linear regression baseline following the USDA Economic Research Service (ERS) yield modeling framework \citep{WestcottJewison2013}. This approach uses agronomically-motivated features (temperature and precipitation during critical growth periods).

The OLS model predicts yield as:
\begin{align}
\hat{y}_t =& \beta_0 + \beta_1 \cdot \text{Trend}_t + \beta_2 \cdot \bar{T}_{\text{crit}} + \beta_3 \cdot \bar{P}_{\text{crit}} \nonumber\\
&+ \beta_4 \cdot \bar{P}_{\text{crit}}^2 + \beta_5 \cdot S_{\text{June}} + \beta_6 \cdot \bar{y}_{\text{past}}
\label{eq:ols_model}
\end{align}
where $\text{Trend}_t = (t - 1970)/100$ captures technological progress, $\bar{T}_{\text{crit}}$ and $\bar{P}_{\text{crit}}$ represent mean temperature and precipitation during critical growth periods (July for corn weeks 26–30, July-August for soybean weeks 26–34), $\bar{P}_{\text{crit}}^2$ captures asymmetric moisture effects, $S_{\text{June}}$ is June precipitation deficit, and $\bar{y}_{\text{past}}$ is 5-year historical yield mean.

    \begin{table*}[htbp]
    \centering
    \begin{tabular}{p{0.4\textwidth}ll}
    \toprule
    \textbf{Measurement Name} & \textbf{Symbol}           & \textbf{Unit}               \\\midrule
    Temperature at 2 Meters                                       & T2M                       & $^\circ$C                   \\ 
    Temperature at 2 Meters Maximum                               & T2M\_MAX                   & $^\circ$C                   \\ 
    Temperature at 2 Meters Minimum                               & T2M\_MIN                   & $^\circ$C                   \\ 
    Wind Direction at 2 Meters                                    & WD2M                      & Degrees                     \\ 
    Wind Speed at 2 Meters                                        & WS2M                      & m/s                         \\ 
    Surface Pressure                                              & PS                        & kPa                         \\ 
    Specific Humidity at 2 Meters                                 & QV2M                      & g/Kg                        \\ 
    Precipitation Corrected                                       & PRECTOTCORR               & mm/day                      \\ 
    All Sky Surface Shortwave Downward Irradiance                 & ALLSKY\_SFC\_SW\_DWN       & MJ/m$^2$/day                \\ 
    Evapotranspiration Energy Flux                                & EVPTRNS                   & MJ/m$^2$/day                \\ 
    Profile Soil Moisture (0 to 1)                                & GWETPROF                  & 0 to 1                      \\ 
    Snow Depth                                                    & SNODP                     & cm                          \\ 
    Dew/Frost Point at 2 Meters                                   & T2MDEW                    & $^\circ$C                   \\ 
    Cloud Amount                                                  & CLOUD\_AMT                & 0 to 1                      \\ 
    Evaporation Land                                              & EVLAND                    & kg/m$^2$/s $\times 10^6$    \\ 
    Wet Bulb Temperature at 2 Meters                              & T2MWET                    & $^\circ$C                   \\ 
    Land Snowcover Fraction                                       & FRSNO                     & 0 to 1                      \\ 
    All Sky Surface Longwave Downward Irradiance                  & ALLSKY\_SFC\_LW\_DWN       & MJ/m$^2$/day                \\ 
    All Sky Surface PAR Total                                     & ALLSKY\_SFC\_PAR\_TOT      & MJ/m$^2$/day                \\ 
    All Sky Surface Albedo                                        & ALLSKY\_SRF\_ALB           & 0 to 1                      \\ 
    Precipitable Water                                            & PW                        & cm                          \\ 
    Surface Roughness                                             & Z0M                       & m                           \\ 
    Surface Air Density                                           & RHOA                      & kg/m$^3$                    \\ 
    Relative Humidity at 2 Meters                                 & RH2M                      & 0 to 1                      \\ 
    Cooling Degree Days Above 18.3 C                              & CDD18\_3                   & days                        \\ 
    Heating Degree Days Below 18.3 C                              & HDD18\_3                   & days                        \\ 
    Total Column Ozone                                            & TO3                       & Dobson units                \\ 
    Aerosol Optical Depth 55                                      & AOD\_55                    & 0 to 1                      \\ 
    Reference Evapotranspiration                                  & ET0                       & mm/day                      \\ 
    Vapor Pressure                                                & VAP                       & kPa                          \\ 
    Vapor Pressure Deficit                                        & VAD                       & kPa                          \\\bottomrule
    \end{tabular}
    \caption{Descriptions of the 31 weather measurements and their units.}
    \label{tab:weather_variables}
    \end{table*}

OLS achieves only 0.227 R² for corn and 0.460 R² for soybean on extreme years (Table~\ref{tab:main_results}), demonstrating that linear assumptions and hand-crafted features fundamentally fail on unprecedented weather events. While the ERS framework was designed for stable conditions \citep{WestcottJewison2013}, extreme weather requires capturing complex nonlinear interactions. VITA achieves 3.2× improvement over OLS for corn (0.729 vs 0.227 R²), confirming that learned representations are essential for climate-resilient yield prediction.

\subsubsection{Lead Time Performance}
\label{sec:lead_time}

For operational forecasting, growers and insurers need yield estimates before harvest. Table~\ref{tab:lead_time} (main text) evaluates performance with end-of-July cutoff: 6 complete years (312 weeks) + 30 weeks of the target year, totaling 342 of 364 weeks (removing the final 22 weeks of August-December weather). This timing is critical in the US Corn Belt, as corn enters late grain fill and soybeans approach reproductive peak \citep{WestcottJewison2013}.

VITA-Sinusoidal achieves 0.689 R² for corn and 0.560 R² for soybean with the July cutoff. Corn performance drops modestly (0.040 R²) while soybean drops substantially (0.162 R²)—a biologically meaningful difference since August is the critical pod-filling period when weather directly determines soybean seed weight \citep{WestcottJewison2013}. No weather forecasts were used; predictions rely solely on observed weather through July.

\begin{table}[htb]
\centering
\begin{tabular}{lccc}
\toprule
\textbf{Crop} & \textbf{VITA Prior} & \textbf{7 Years} & \textbf{5 Years} \\
& & R² (RMSE) & R² (RMSE) \\
\midrule
\multirow{2}{*}{Corn} & Std. Normal & 0.706 (17.1) & 0.666 (18.2) \\
& Sinusoidal & 0.729 (16.3) & 0.697 (17.4) \\
\midrule
\multirow{2}{*}{Soybean} & Std. Normal & 0.698 (5.2) & 0.694 (5.3) \\
& Sinusoidal & 0.722 (5.0) & 0.684 (5.3) \\
\bottomrule
\end{tabular}
\caption{Reduced temporal context on extreme years: 5 years (260 weeks) vs 7 years (364 weeks) of weather data.}
\label{tab:reduced_context}
\end{table}

VITA-Sinusoidal still achieves 3.0× improvement over OLS for corn (0.689 vs 0.227) and 1.5× for soybean (0.560 vs 0.382) despite identical cutoffs. Importantly, VITA-Sinusoidal shows superior robustness: while SimMTM drops 0.096 R² (corn) and 0.204 R² (soybean), and T-BERT drops 0.071 R² (corn) and 0.185 R² (soybean), VITA-Sinusoidal degrades less (0.040 R² corn, 0.162 R² soybean), confirming its viability for pre-harvest operational forecasting.

\subsubsection{Reduced Temporal Context}
\label{sec:reduced_context}

To test data efficiency, we evaluate extreme years using only 5 years of weather history (260 weeks) instead of 7 years (364 weeks)—a 28\% reduction with practical implications for data-scarce regions or faster inference. Table~\ref{tab:reduced_context} shows VITA-Sinusoidal achieves 0.697 R² for corn and 0.684 R² for soybean, outperforming most baselines in Table~\ref{tab:main_results} despite limited context.

Performance degradation is modest: corn drops 0.032 R² (sinusoidal) and soybean 0.038 R². The standard normal prior shows even greater robustness (0.040 R² drop for corn, 0.004 R² for soybean), indicating both variants effectively leverage available temporal information when context is limited.

\bibliography{references}

\end{document}